%% file: neurips_2026.tex
\documentclass{article}

\usepackage[eandd, final]{neurips_2026}

\usepackage[utf8]{inputenc} %
\usepackage[T1]{fontenc}    %
\usepackage{amsmath}
\usepackage{hyperref}       %
\usepackage{url}            %
\usepackage{booktabs}       %
\usepackage{amsfonts}       %
\usepackage{nicefrac}       %
\usepackage[protrusion=true,expansion=false]{microtype}
\usepackage{xcolor}         %
\usepackage{graphicx} %
\usepackage{comment} %
\usepackage{makecell} %
\usepackage{wrapfig} %
\usepackage{enumitem} %
\usepackage{float}
\usepackage{multicol}
\usepackage{multirow}
\usepackage{subcaption}
\usepackage{colortbl} %
\definecolor{LightGreen}{HTML}{D8F5D0} %
\definecolor{LightRed}{HTML}{F8D7DA} %
\usepackage{tikz}%

\definecolor{GroupA}{HTML}{482878}
\definecolor{GroupB}{HTML}{E30280}

\newcommand{\circnumA}[1]{%
\tikz[baseline=(char.base)]{
    \node[
        rounded corners=6pt,
        fill=GroupA,
        text=white,
        inner xsep=5pt,
        inner ysep=2pt,
        font=\bfseries\scriptsize
    ] (char) {#1};
}}

\newcommand{\circnumB}[1]{%
\tikz[baseline=(char.base)]{
    \node[
        rounded corners=6pt,
        fill=GroupB,
        text=white,
        inner xsep=5pt,
        inner ysep=2pt,
        font=\bfseries\scriptsize
    ] (char) {#1};
}}

\title{What Survival Benchmarks Don’t Tell You:\newline Impact of Model Selection and Dataset Regimes}

\author{%
  Sandra Dening\textsuperscript{1}\thanks{Equal contribution.}
  \qquad
  Emre Kavak\textsuperscript{1,2,3}\footnotemark[1]
  \qquad
  Christian Wachinger\textsuperscript{1,2,3}
  \qquad
  George H. Chen\textsuperscript{4}
  \\[0.6em]
  \textsuperscript{1}Technical University of Munich, Germany
  \\
  \textsuperscript{2}Konrad Zuse School of Excellence in Reliable AI (relAI), Germany
  \\
  \textsuperscript{3}Munich Center for Machine Learning (MCML), Germany
  \\
  \textsuperscript{4}Heinz College, Carnegie Mellon University, USA
  \\[0.6em]
  \texttt{sandra.dening@tum.de}
  \quad
  \texttt{emre.kavak@tum.de}
  \\
  \texttt{christian.wachinger@tum.de}
  \quad
  \texttt{georgechen@cmu.edu}
}

\counterwithin{figure}{section}
\counterwithin{table}{section}

\begin{document}

\maketitle

\begin{abstract}\vspace{-1em}
Despite the proliferation of survival analysis methods, current benchmarks often operate under narrow conditions: they heavily rely on the concordance index (C-index) for evaluation and model selection, and frequently either exclude deep learning (DL) methods or underrepresent challenging dataset regimes. To address these limitations, we conduct a large-scale, neutral benchmark comparing 10 classical, tree-based, and DL methods across 49 diverse datasets. Crucially, we treat the model selection criterion (optimizing for C-index, Integrated Brier Score (IBS), or Mean Absolute Error (MAE-PO)) as an independent experimental variable. Using Bayesian mixed-effects modeling, we demonstrate that evaluation design heavily dictates conclusions. For complex DL and tree-based architectures, selecting hyperparameters to optimize for IBS instead of C-index yields substantial improvements in probabilistic accuracy and calibration at a negligible cost to discriminative ranking. Furthermore, method performance (C-index, IBS, MAE-PO, and X-CAL) is fundamentally governed by dataset characteristics. While DL models scale exceptionally well with larger sample sizes, they suffer notable performance drops in high-dimensional feature spaces. Additionally, we expose a structural vulnerability where heavy censoring artificially inflates absolute calibration metrics, creating overly conservative tests.  Ultimately, our findings challenge the utility of global single-number performance summaries and advocate for survival analysis results that transparently report the evaluation setup. Our code is available at: \url{https://github.com/sdening/surv-bench-study}
\end{abstract}

\section{Introduction}

The Cox proportional hazards model \citep{Cox1972RegressionModels} has been a cornerstone method of survival analysis (time-to-event modeling) for over fifty years. While regularized extensions of it such as CoxNet \citep{Simon2011RegularizationPathsCox} remain competitive today, the methodological landscape has rapidly expanded with, for instance, tree-based ensembles (e.g., \citep{Ishwaran2008RandomSurvivalForests,alberge2025survival}) and a recent wave of deep survival models (e.g., \citep{Katzman2018DeepSurv, Lee2018DeepHit, kvamme2019timetoeventpredictionneuralnetworks, survtrace_2022_wang, Knottenbelt_2025, Wiegrebe_2024}). Yet, the evaluation practices used to compare these methods have not kept pace. Results are typically summarized through average ranks, winner counts on Harrell's concordance index \citep{harrell1982evaluating} (or variants of it such as those by \citet{Antolini2005TimeDependentCindex} and \citet{uno2011c}; for ease of exposition, throughout the paper we abbreviate these collectively simply as the ``C-index'', specifying a particular variant as needed) or mean metric values across datasets---a compression that obscures crucial method-level variation. This paper investigates what this standard evaluation process misses, demonstrating that the field has underestimated the nuances of model performance across different dataset regimes and especially across different model selection criteria.

Current neutral benchmarks for survival analysis generally converge on a stable consensus: no modern method consistently outperforms classical Cox-based methods \citep{chien2024benchmarking, burk2026largescaleneutralcomparisonstudy, Herrmann_2020_largescaleBM_SA_multiOmics, Birolo_2025_beyondCox, Zhang2022SurvBenchmark}. However, this conclusion is drawn under evaluation conditions that leave large areas of the methodological space unexamined. As highlighted by \citet{lillelund2025stopchasingcindexevaluate}, survival analysis model evaluations overwhelmingly rely on the C-index, neglecting probabilistic accuracy, time-to-event accuracy, and calibration. Crucially, the \textit{model selection strategy} (e.g., the choice of evaluation metric driving hyperparameter and checkpoint selection) is almost never varied. Papers introducing new models frequently select hyperparameters based on the C-index without exploring how other hyperparameter selection strategies might have performed (e.g., \citep{Katzman2018DeepSurv, Lee2018DeepHit, kim2026tabularfoundationmodelssurvival}). Taken together, the conclusion that Cox remains competitive may hold under these narrow conditions, including datasets of small sample and feature size, but it has not been robustly tested in regimes where alternatives could plausibly excel. 

To address this, we conduct a large-scale benchmark treating evaluation conventions as experimental variables. We compare 10 methods (across linear Cox, tree-based, and deep learning method families) across 49 diverse datasets, each independently optimized under three model selection criteria on a validation set: time-dependent Antolini C-index ($C^{td}$) \citep{Antolini2005TimeDependentCindex}, Integrated Brier Score (IBS) \citep{Graf1999}, and a Pseudo-Observation-based Mean Absolute Error (MAE-PO) \citep{qi2023effectivemeaningfulwayevaluate}. On the held-out test set, all three metrics plus calibration (measured via X-CAL \citep{goldstein2021_XCal}) are computed for every experiment, yielding a comprehensive set of results across methods, datasets, and selection strategies. Using Bayesian Generalized Linear Mixed-effects Models (GLMMs) \citep{gelman2007data}, we estimate effect strengths of different components in our benchmark on target metrics, thereby accounting for interaction and confounding effects. We further complement all these findings with descriptive group-wise mean and median trends in the Appendix (\ref{app:suppl_material}). Our core findings fall into two main categories: 

\noindent\textbf{\circnumA{A} The Evaluation vs. Optimization Gap: Moving Beyond C-index Selection}
\begin{enumerate}[leftmargin=*, widest=\circnumA{A.3}, labelsep=0.6em, topsep=2pt, itemsep=1pt, parsep=0pt]
    \item[\circnumA{A.1}] \textbf{What Drives Performance Depends on the Metric:} C-index typically varies little between methods within a dataset, but much more across datasets. Other metrics tell a different story: Probabilistic accuracy (IBS) and calibration (X-CAL), in contrary to C-index, show significant sensitivity to optimization and algorithmic choice.
    \item[\circnumA{A.2}] \textbf{The Benefits of Proper Scoring Rules as Model Selection Criteria:} Relying solely on the C-index for hyperparameter selection fails to regularize the predictive distributions of complex deep learning and tree-based methods.  Shifting model selection to optimize a proper scoring rule like the Integrated Brier Score (IBS) yields significant gains in probabilistic accuracy and calibration, without significantly altering the model's discriminative ranking.
    \item[\circnumA{A.3}] \textbf{Time-to-event Accuracy Requires Direct Optimization:} When models are tuned by optimizing C-index or IBS, their ability to improve Mean Absolute Error (MAE-PO) over a simple Kaplan-Meier (KM) baseline \citep{Kaplan1958NonparametricEstimation} is essentially a coin flip (50\%).  By instead tuning models to optimize MAE-PO, the models we benchmark outperform the KM baseline 72\% of the time while reducing extreme temporal prediction outliers. 
\end{enumerate}
\vspace{0.5em}
\noindent\textbf{\circnumB{B} Contextualizing Performance: The Role of Dataset Regimes}
\vspace{-0.3em}
\begin{enumerate}[leftmargin=*, widest=\circnumB{B.2}, labelsep=0.6em, topsep=2pt, itemsep=1pt, parsep=0pt]
    \item[\circnumB{B.1}] \textbf{Dataset Scale and Dimensionality Govern Method Performance:} Tree-based and deep learning methods consistently scale to larger sample sizes. However, deep learning models struggle with high-dimensional (feature count ($P$) > observation count ($N$)) data, suffering notable IBS and MAE performance drops as feature counts increase.
    This hints that deep learning may be overfitting and may benefit from regularization strategies when many features are available.
    \item[\circnumB{B.2}] \textbf{The Calibration Fallacy: Relative vs Absolute Utility.} We observe artificial improvement in X-CAL scores as the dataset censoring rate increases. While X-CAL remains a robust and highly informative metric for the \textit{relative} ranking of models (i.e., comparing one architecture against another on the same data), its \textit{absolute} value becomes highly unreliable under heavy censoring. Consequently, standard D-Calibration tests and X-CAL objectives become overly conservative, masking underlying miscalibration and falsely labeling models as well-calibrated simply because they are evaluated on highly censored datasets.
\end{enumerate}

\section{Related Work}
\label{sec:related_work}
Survival analysis benchmarking has matured considerably in recent years, yet several structural gaps continue to limit the conclusions these studies can draw. We summarize eight dedicated benchmarks in Tab.~\ref{tab:benchmark_summary}; these are studies that compare existing methods without proposing new ones. Despite the progress in scale and evaluation diversity, four interrelated gaps persist: systematic underrepresentation of deep learning methods, limited dataset diversity, inconsistent evaluation methodology, and the treatment of model selection criteria as a fixed convention rather than a design variable.

\textbf{Deep Learning is Systematically Underrepresented.}
Most benchmarks in Tab.~\ref{tab:benchmark_summary} are R-based, structurally excluding recent, Python-native DL methods despite \citet{Wiegrebe_2024} cataloging over 60 DL survival methods when they wrote their review. \citet{Herrmann_2020_largescaleBM_SA_multiOmics} and \citet{burk2026largescaleneutralcomparisonstudy} include no DL methods; \citet{burk2026largescaleneutralcomparisonstudy} explicitly cite unreliable integration of DL methods into their R-based benchmark suite as the reason for their exclusion. \citet{Zhang2022SurvBenchmark} include three DL methods but without tuning. \citet{Birolo_2025_beyondCox} evaluate tuned DL methods on six mid-sized biomedical datasets and their study is the only benchmark to report Antolini's time-dependent C-index \citep{Antolini2005TimeDependentCindex}, which they show is critical for fair evaluation of non-proportional-hazards methods, yet no other benchmark in Tab. \ref{tab:benchmark_summary} adopts it for evaluation.

\textbf{Dataset Diversity is Limited in Both Domain and Scale.} \citet{burk2026largescaleneutralcomparisonstudy} provide perhaps the most comprehensive method comparison to date with 32 datasets across multiple domains and two model selection criteria, but restrict their study to low-dimensional settings ($N>P$) and exclude DL. \citet{Herrmann_2020_largescaleBM_SA_multiOmics} cover high-dimensional omics data but include no DL. \citet{Birolo_2025_beyondCox} include tuned DL methods but are limited to six small-to-medium ($N>P$) biomedical datasets. \citet{green2026finsurvivalsuitelargescale} are the first to leave the biomedical domain entirely and provide by far the largest datasets in any survival benchmark, with 16 tasks containing up to millions of observations. While a valuable contribution to large-scale evaluation, all tasks share the same data-generating process and censoring mechanism, limiting generalizability, and DL methods are evaluated without a systematic tuning protocol and inconsistently on different variants of the C-index -- conditions under which DL exhibited particularly poor performance that the authors themselves attribute to tuning difficulties. No existing benchmark simultaneously spans multiple domains, covers both low- and high-dimensional regimes, and includes systematically tuned DL methods.

\textbf{Evaluation Methodology is Inconsistent and Incomplete.} Harrell's C-index \citep{harrell1982evaluating} --- known to be inappropriate for non-proportional-hazards models \citep{Birolo_2025_beyondCox} --- remains the primary evaluation metric in most studies. Moreover, benchmarks disagree on which C-index variant to use: Harrell's, Uno's \citep{uno2011c}, Antolini's \citep{Antolini2005TimeDependentCindex}, or other modified versions. Those C-index variants are known to yield different rankings on identical data \citep{sierra2025cindexmultiverse}. As \citet{lillelund2025stopchasingcindexevaluate} observe, the research community remains focused on optimizing the C-index, and \citet{zhou_2023} find that it serves as the primary performance measure for over 80\% of survival analysis papers. This focus continues even though the metric has well-known limitations for non-proportional-hazards models, cannot assess predictive accuracy and offers limited practical usefulness -- and despite \citet{Herrmann_2020_largescaleBM_SA_multiOmics} recommending already in 2020 that IBS ``should be considered more important'' given that C-index is not a strictly proper scoring rule. Calibration is absent from several benchmarks entirely. Some benchmarks claim to assess calibration via IBS, but as \citet{chen2024introductiondeepsurvivalanalysis} states and we show, calibration needs a distinct focus. No prior benchmark includes a time-accuracy metric, despite \citet{qi2023effectivemeaningfulwayevaluate} establishing MAE-PO as a proper scoring rule for this purpose. To our knowledge, however, no benchmark has incorporated it at scale, even though accurate event-time estimation is highly relevant in many real-world applications.

\textbf{Model Selection Criteria are Rarely Varied.} Finally, model selection criteria have been treated as a convention rather than a variable. Only \citet{burk2026largescaleneutralcomparisonstudy} experiment with two selection targets, but do not analyze how the choice affects rankings or calibration. No benchmark investigates the interaction between the model selection criteria and the target evaluation metric, which we show is a consequential evaluation design decision. 

A consistent substantive finding across all eight benchmarks is that no single method dominates. Notably, \citet{Herrmann_2020_largescaleBM_SA_multiOmics} observed already that method rankings change between C-index and IBS --- yet no subsequent benchmark has systematically investigated this metric dependence. 
More consequentially, most studies conclude that the classical (regularized) Cox Proportional Hazards model remains competitive with modern alternatives and recommend it as the practical default -- a conclusion we show holds only under specific evaluation and dataset characteristic regimes that prior benchmarks have often implicitly assumed.

\section{Benchmark and Experimental Setup}
\label{experimental_setup}
\subsection{Datasets}
In total, our benchmark comprises 49 survival analysis datasets (Tab. \ref{tab:dataset_overview}). These were deliberately selected to span a wide range of application domains (clinical, omics, engineering, socio-economics, etc.), sample sizes (26 to $100{,}000$), feature sizes (3 to $15{,}500$), censoring rates (0\% to >95\%), and time-to-event distributions. The objective is not to evaluate models on a small set of commonly used datasets, but to assess their performance across a sufficiently diverse landscape that reveals strengths and weaknesses under varying data regimes.

We group the datasets into two tiers. Tier 1 comprises 43 datasets from the SurvSet repository \citep{drysdale2022survsetopensourcetimetoeventdataset}. Tier 2 adds six datasets from external sources to cover domains and scales underrepresented in SurvSet and to include datasets widely used as reference points in the survival analysis literature: SUPPORT2 \citep{Knaus1995SUPPORT}, METABRIC \citep{Curtis_2012_metabric, Pereira_2016_metabric}, KKBox (subsampled to $N=100{,}000$) \citep{kvamme2019timetoeventpredictionneuralnetworks, kvamme2019brierscoreadministrativecensoring}, MIMIC-IV \citep{Johnson2023_mimiciv}, ADNI \citep{Petersen2010ADNI}, and a custom-processed financial dataset from the FinSurvival suite \citep{green2026finsurvivalsuitelargescale}.

\textbf{Selection Criteria.} Datasets are included if they contain a continuous time variable with a binary right-censoring indicator, at least two predictive features, static tabular data, and are publicly available (with or without an application process). Unlike \citet{burk2026largescaleneutralcomparisonstudy}, we impose no minimum event count or dimensionality restriction, deliberately retaining small sample ($N = 26$), high-censoring (up to 95\%), and high-dimensional ($P>N$) regimes. Datasets with a sample size exceeding $N = 100{,}000$ are subsampled with stratification on the event indicator as very large datasets are impractical to run on our standardized and unified training pipeline. The 49 datasets satisfy the minimum of $43$ recommended by \citet{Boulesteix2017EvidenceBasedBenchmarking} for sufficient statistical power.

\textbf{Preprocessing.} In the preprocessing, observations with non-positive follow-up time are removed. All datasets pass through a unified pipeline: (1) category summarization (collapsing levels with more than 10 categories into "other"), (2) one-hot-encoding with reference-category dropping, (3) z-score standardization of continuous features, (4) listwise deletion of remaining missing values, (5) near-zero variance removal (numeric: $\text{standard deviation} < 10^{-5}$; one-hot: prevalence $< \max (0.0025, 5/N)$), and (6) high-correlation removal (Spearman $\rho >0.98$). Tier 2 datasets are not available through SurvSet and require dataset-specific loading and cleaning before entering the shared pipeline. Where an established reference preprocessing exists, we follow it to preserve comparability with prior work. For more details on Tier-2 preprocessing see Appendix \ref{app:dataset_preprocessing}. 

\subsection{Methods}
\label{sec:methods_relatedwork}
We benchmark 10 time-to-event prediction methods spanning three method families. Comprehensive implementation details, including software libraries, model classes, and exhaustive hyperparameter search spaces, are provided in Appendix~\ref{app:methods_and_imp} and corresponding Tabs. \ref{tab:method_overview}, \ref{tab:hp_grids}.

\textbf{Linear Cox (2 methods).} We evaluate the original unregularized Cox proportional hazards baseline without hyperparameter tuning (abbreviated CoxPH) \citep{Cox1972RegressionModels}, and CoxNet \citep{Simon2011RegularizationPathsCox}, an elastic net regularized Cox model.

\textbf{Tree-Based (3 methods).} We evaluate Random Survival Forest (RSF) \citep{Ishwaran2008RandomSurvivalForests} and Gradient Boosting Survival Analysis (GrBoost) \citep{Friedman2001GradientBoosting}. We additionally benchmark XGBoost \citep{Chen_2016_XGboost}, training it with both Cox and Accelerated Failure Time (AFT) objectives. Survival functions for XGBoost are obtained via the Breslow estimator or a closed-form AFT construction (Appendix \ref{app:xgboost}).

\textbf{Deep Learning (5 methods).} DeepSurv uses a single-hidden-layer MLP trained with the Cox partial log-likelihood and constructs survival curves via the Breslow estimator \citep{Katzman2018DeepSurv}. DeepHit \citep{Lee2018DeepHit} is a discrete-time model whose loss combines a discrete-time log-likelihood with a ranking loss term, trained across varying temporal discretization schemes. CoxTime \citep{kvamme2019timetoeventpredictionneuralnetworks} extends the Cox framework by making the network a function of both covariates and time. SurvTRACE \citep{survtrace_2022_wang} applies a BERT-style transformer over tabular features, utilizing quantile-based time discretization; it is trained from scratch using a multi-task learning framework that incorporates auxiliary mortality and ``length-of-stay'' (i.e., survival time) prediction tasks to learn robust shared representations. Finally, CoxKAN \citep{Knottenbelt_2025} replaces the standard MLP in DeepSurv with a Kolmogorov-Arnold Network to produce interpretable symbolic expressions for the log-partial hazard.

\subsection{Evaluation}
\paragraph{Training Protocol.} 
\label{sec:training_protocol}
\setlength{\columnsep}{5pt}
\begin{wrapfigure}[14]{r}[1cm]{0.60\textwidth}
    \vspace{-2.5em}
    \hspace*{-0.5cm}
    \centering
    \includegraphics[width=\linewidth]{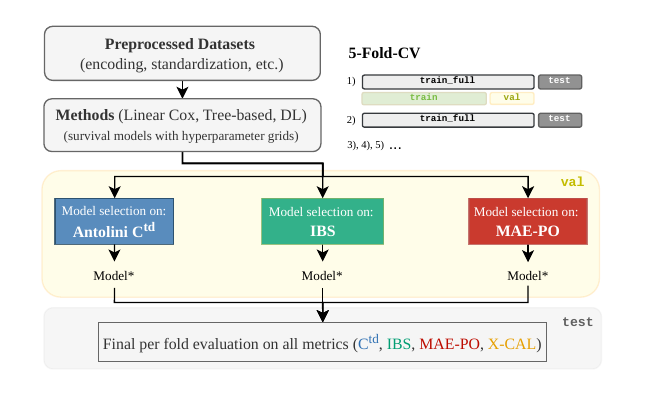}
    \caption{Benchmark setup overview.}
    \label{fig:benchmark_setup_overview}
    \vspace{-0.0em}
\end{wrapfigure}
Fig.~\ref{fig:benchmark_setup_overview} shows the setup: each dataset is evaluated using 5-fold cross-validation, stratifying on the binary event indicator. Within each fold, the training set is further split into 80/20 train and validation subsets. Hyperparameters are selected via exhaustive grid search on the validation set to ensure exact reproducibility given the same random seed. Linear Cox and tree-based methods without an internal notion of training progression are refit on the full train+validation set before test evaluation after hyperparameter selection. In contrast, deep learning models and XGBoost include an additional selection step over epochs or boosting rounds, performed via validation-based early stopping (with a patience of 10 epochs/rounds). The resulting checkpoint is used directly for test evaluation without refitting.

A compute budget of 15 hours per experiment is enforced to ensure tractability of the large-scale exhaustive grid search across all $10 \times 49 \times 3 = 1{,}470$ experiments. This threshold was manually chosen empirically from initial runs across datasets of varying sizes. All experiments are run on CPUs (see Appendix~\ref{app:reproducibility} for details), as GPU acceleration yielded negligible gains due to evaluation metric computations (C-index, IBS, MAE-PO) being dominated by \texttt{NumPy}/\texttt{pandas} CPU operations.

\paragraph{Triple Model Selection Criteria.} Every experiment is evaluated under three distinct validation-based model selection criteria: $C^{td}$, IBS, and MAE-PO. Using these three different targets is central to the evaluation design analysis in Section \ref{sec:benchmarking_results}: it enables direct comparison of how the same method behaves when its hyperparameters are tuned to optimize different evaluation criteria, isolating the effect of the evaluation design choice from the effect of the method itself. 

\paragraph{Evaluation Metrics.} On the held-out test set, four metrics are computed for every method, fold, and model selection criterion. To ensure comparability, all metrics are evaluated on a shared time grid, constructed from training follow-up times and trimmed to avoid IPCW boundary instability. All native survival curves -- which differ substantially across method families -- are mapped to this grid via linear interpolation following \citet{Archetti2024InterpolationSurvival} (full construction details in Appendix \ref{app:eval_time_grid}). The four metrics are: 

\begin{itemize}[leftmargin=*, labelsep=0.5em]
    \item \textit{Discrimination}: Antolini $C^{td}$ (higher is better, abbreviated ``$\uparrow$'') \citep{Antolini2005TimeDependentCindex} as a time-dependent concordance index that does not assume proportional hazards and -- as shown by \citet{Birolo_2025_beyondCox} -- is better suited for non-proportional-hazards methods. We compute $C^{td}$ scores via \texttt{pycox}'s \texttt{EvalSurv} Python class \citep{kvamme2019timetoeventpredictionneuralnetworks} with the \texttt{adj\_antolini} tie-handling.
    \item \textit{Probabilistic Accuracy}: Integrated Brier Score (IBS) (lower is better, $\downarrow$) with IPCW weighting \citep{Graf1999}, computed via \texttt{SurvivalEVAL} \citep{qi2024survivaleval}. 
    As IBS is dataset dependent, we also report $\text{IBS}_{\text{rel}} = (1-\text{IBS}_{\text{method}}/\text{IBS}_{\text{KM}}) \cdot 100$ (where $\text{IBS}_{\text{method}}$ refers to the IBS score for a method being evaluated, and $\text{IBS}_{\text{KM}}$ refers to the IBS score achieved by the Kaplan-Meier baseline estimator), measuring percentage improvement over the KM baseline. 
    \item \textit{Time Accuracy}: MAE-PO ($\downarrow$) -- mean absolute error using pseudo-observations to handle censoring \citep{qi2023effectivemeaningfulwayevaluate}, computed via \texttt{SurvivalEVAL} \citep{qi2024survivaleval}; shown by \citet{qi2023effectivemeaningfulwayevaluate} to best approximate the true MAE across a variety of censoring distributions. We report $\text{MAE}_{\text{rel}} = (1- \text{MAE}_{\text{method}}/\text{MAE}_{\text{KM}}) \cdot 100$, analogously to $\text{IBS}_{\text{rel}}$, and $\text{MAE}_{\text{ratio}} = \text{MAE}_{method}/\text{MAE}_{\text{KM}}$.
    \item \textit{Calibration}: X-CAL ($\downarrow$) \citep{goldstein2021_XCal} -- a continuous, differentiable relaxation of distributional calibration (D-Calibration) \citep{Haider2020SurvivalDistributions} that measures calibration via the probability integral transform. Calibration is also computed using \texttt{SurvivalEVAL} \citep{qi2024survivaleval}. While D-Calibration yields a binary hypothesis test (calibrated or not), X-CAL provides a continuous score suitable for comparison across methods.
\end{itemize}

\section{Benchmarking Results}
\label{sec:benchmarking_results}

\subsection{Statistical Evaluation: Bayesian Mixed-Effects Modeling}
To rigorously disentangle the performance drivers in our benchmark, we evaluate the results using a Bayesian Generalized Linear Mixed-effects Modeling (GLMM) framework. Simple aggregate rankings often confound an algorithm's true capability with the baseline difficulty of the datasets it was evaluated on. To correct for this, we model the predictive performance as a function of fixed effects: the learning algorithm, the model selection strategy, dataset meta-features (e.g., dataset observation and feature count, and censoring rate), and their interactions, while utilizing a random intercept for each dataset to absorb intrinsic, unobserved dataset heterogeneity. 
We have one model for each of IBS, C-index, MAE-PO (ratio), and X-CAL. The governing formula for the linear predictor across all our GLMMs reads:
\begin{equation}
    \eta_{ijk} = \beta_0 + \alpha_i + \gamma_j + (\alpha\gamma)_{ij} + \mathbf{m}_k^\top \boldsymbol{\delta} + \mathbf{m}_k^\top \boldsymbol{\theta}_i + u_k
\end{equation}
where:
\begin{itemize}
    \item $\eta_{ijk}$ represents the linear predictor for method $i$, model selection strategy $j$, and dataset $k$ (averaged across the 5 cross-validation folds), mapped to the conditional mean $\mu_{ijk}$ via an identity link (Gaussian for $C^{td}$ and IBS) or a log link (Gamma for X-CAL and MAE-PO ratio).
    \item $\beta_{0}$ is the global intercept.
    \item $\alpha_{i}$ and $\gamma_{j}$ define the fixed main effects for the learning method and the model selection strategy, respectively, with $(\alpha\gamma)_{ij}$ capturing their two-way interaction.
    \item $m_{k} \in \mathbb{R}^4$ is a vector of standardized dataset meta-features: log-transformed sample size ($\log N$), log-transformed feature count ($\log P$), log-transformed time range ($\log T$), and the censoring rate.
    \item $\delta \in \mathbb{R}^4$ is a vector representing the fixed main effects of these meta-features.
    \item $\theta_{i} \in \mathbb{R}^4$ captures the interaction between method $i$ and the dataset meta-features, quantifying how specific architectures scale with sample size, dimensionality, or censoring.
    \item $u_{k} \sim \mathcal{N}(0,\sigma_{u}^{2})$ is a random intercept for dataset $k$, absorbing unobserved dataset-specific baseline difficulty.
\end{itemize}

Models were fitted using Hamiltonian Monte Carlo via the \texttt{bambi} \citep{Capretto2022} and \texttt{PyMC} \citep{pymc2023} libraries. This fully probabilistic approach allows us to report the Probability of Direction ($pd$) \citep{makowski2019indices} for specific performance gains and to perform variance partitioning: estimating the relative contributions of combined fixed effects, dataset-level random effects, and residual variation to modeled performance variance. Probability of Direction ($pd$), posterior predictive checks, and the variance decomposition underlying Fig. \ref{fig:combined_variance_partitioning} are detailed in Appendices \ref{app:probability_of_direction}, \ref{app:posterior_analysis}, and \ref{app:variance_decomposition}, respectively.

\subsection{Run Completion and Failure Analysis} %
At the scale of this benchmark, incomplete runs are expected: not all 1{,}470 experiments completed successfully. Our main analyses are performed on the unbalanced union of successful runs. %

We mark a dataset-method pair as failed if any fold did not produce results. The failures are not random but reflect interpretable scalability limits of specific architectures: mathematical errors (e.g. unregularised CoxPH on ill-conditioned design matrices), compute budget timeouts (e.g. RSF, GrBoost, CoxKAN), and memory exhaustion (e.g. CoxTime, SurvTRACE). Failure counts also differ marginally across selection criteria ($C^{td}$: 48,  IBS: 50, MAE-PO: 56), as some validation metrics are more expensive to compute. Appendix \ref{app:failure_analysis} provides a detailed breakdown of runtime, memory, and numerical failures, distinguishing timeouts shared across all model-selection criteria from those specific to individual criteria. Figure~\ref{fig:failure_rate_matrix} illustrates these systematic patterns for $C^{td}$-based model selection, showing how failures concentrate in specific method–regime combinations.

To verify that the systematic failures do not harm our analysis methods through survivorship bias, especially the Bayesian GLMM, we also conducted the same analysis on the common intersection of successful runs, restricting the analysis to datasets on which all methods completed successfully. The overarching trends and main findings remain completely unchanged and we conclude that our results are robust in these terms. Detailed results are provided in Appendix \ref{app:robustness_common_intersection}, Fig \ref{fig:pd_heatmaps_intersection}. 

\subsection{Key Findings}
\label{sec:findings}
The following findings \circnumA{A} and \circnumB{B} should not be interpreted independently. Instead, they form an interdependent evaluation protocol. Practitioners should first determine what they aim to optimize through an application-specific evaluation objective, as this influences the different performance metrics on the test set.

\textbf{\circnumA{A} The Evaluation vs. Optimization Gap: Moving Beyond C-index Selection}

\begin{figure}[H]
    \centering
    \includegraphics[width=0.6\textwidth]{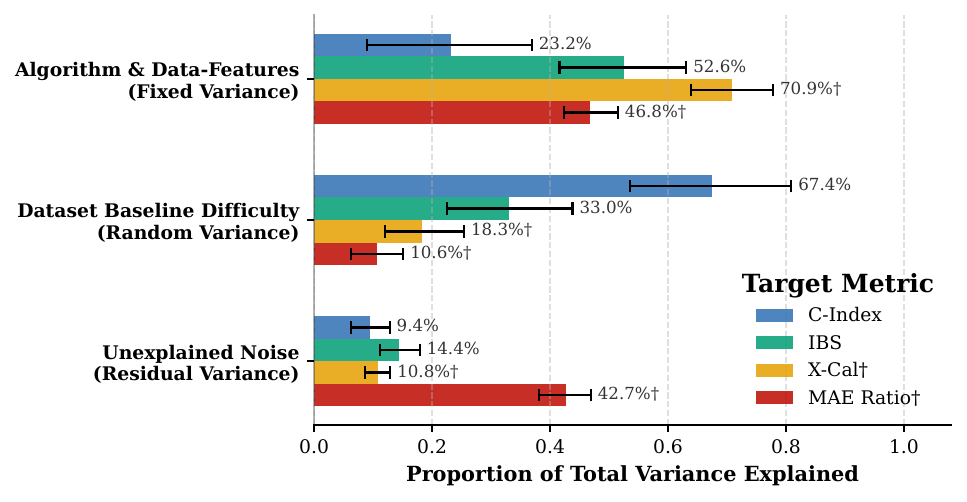}
    \caption{Bayesian Variance Partitioning Across Evaluation Metrics with 94\% highest-density interval. $\dagger$ denotes log scale.}
    \label{fig:combined_variance_partitioning}
\end{figure}

Our Bayesian GLMM analysis reveals a profound dichotomy in how performance variance is explained across different evaluation metrics, demonstrating that standard C-index selection systematically masks the true capabilities of complex architectures (Fig. \ref{fig:combined_variance_partitioning}).

\textbf{\circnumA{A.1} What Drives Performance Depends on the Metric.} For discriminative performance (C-index), a model's performance is largely predetermined by the dataset heterogeneity.  It acts as an overpowering constraint, explaining 67.4\% of the variance, while the combined fixed effects of model choice, selection strategy, and dataset meta-features account for only 23\% (Fig. \ref{fig:combined_variance_partitioning}).  Consequently, for practitioners primarily concerned with ranking, the choice of algorithm seems less important, gaining importance only in large $N$ and large $P$ regimes (see Fig.~\ref{fig:kfold_cindex_heatmap_lowdim}, \ref{fig:kfold_cindex_heatmap_highdim}, and \ref{fig:pd_heatmaps}).

Conversely, probabilistic accuracy (IBS) and calibration (X-CAL) are highly susceptible to algorithmic intervention.  For IBS, our controlled choices explain the majority of the variance (52.6\%), and for X-CAL, this rises dramatically to 70.9\%. This indicates that while discrimination is largely a property of the data, proper calibration and probabilistic scoring are fundamentally properties of the model. Therefore, adopting a targeted, dataset-regime-specific model selection strategy is not merely an optimization detail, but a critical requirement for unlocking substantial performance gains.

\textbf{\circnumA{A.2} The Benefits of Proper Scoring Rules as Model Selection Criteria.}
Our analysis reveals a profound asymmetry in model selection: optimizing for strictly proper scoring rules unlocks significant performance gains for highly flexible models without sacrificing their discriminative ranking. The C-index evaluates concordance but ignores the scale of predicted probabilities, allowing models to generate wildly uncalibrated distributions. In contrast, optimizing hyperparameters for IBS forces models to learn well-calibrated probabilistic distributions, while largely preserving their ranking capabilities in our benchmark.

\begin{table}[t]
\centering
\caption{
Effects of model selection strategy on target metrics, when switching selection strategy $\text{From} \rightarrow \text{To}$. Insignificant results ($pd$ $< 0.90$) are kept colorless. Significant results are color-coded (light green for beneficial, light red for detrimental) and marked with asterisks ($^{*}$ $pd$ $\geq$ 0.90, $^{**}$ $pd$ $\geq$ 0.95, $^{***}$ $pd$ $\geq$ 0.99). A dagger ($\dagger$) denotes metrics that are reported on a log-multiplier scale.
}
\label{tab:selection_switch_effects}
\vspace{0.5em}
\resizebox{\linewidth}{!}{%
\begin{tabular}{lrrrr|rrrr|rrrr}
\toprule
& \multicolumn{4}{c}{$\text{C-index} \rightarrow \text{IBS}$}
& \multicolumn{4}{c}{${\text{C-index}} \rightarrow \text{MAE}$}
& \multicolumn{4}{c}{$\text{IBS} \rightarrow \text{MAE}$} \\
\cmidrule(lr){2-5}
\cmidrule(lr){6-9}
\cmidrule(lr){10-13}
\textbf{Method}
& $\Delta \text{C-index}\uparrow$ & $\Delta \text{IBS}\downarrow$ & $\Delta \text{MAE}_{\mathrm{ratio}}^{\dagger}\downarrow$ & $\Delta \text{X-CAL}^{\dagger}\downarrow$
& $\Delta \text{C-index}\uparrow$ & $\Delta \text{IBS}\downarrow$ & $\Delta \text{MAE}_{\mathrm{ratio}}^{\dagger}\downarrow$ & $\Delta \text{X-CAL}^{\dagger}\downarrow$
& $\Delta \text{C-index}\uparrow$ & $\Delta \text{IBS}\downarrow$ & $\Delta \text{MAE}_{\mathrm{ratio}}^{\dagger}\downarrow$ & $\Delta \text{X-CAL}^{\dagger}\downarrow$ \\
\midrule

coxnet & \cellcolor[HTML]{f8d7da} $-0.009^{*}$ & $-0.002$ & $0.163$ & $-0.157$ & $-0.002$ & $-0.002$ & $-0.420$ & $0.052$ & $0.007$ & $-0.001$ & \cellcolor[HTML]{d4edda} $-0.584^{*}$ & \cellcolor[HTML]{f8d7da} $0.208^{*}$ \\
rsf & $-0.000$ & $-0.001$ & $-0.007$ & $-0.039$ & $-0.001$ & $-0.001$ & $-0.014$ & $-0.057$ & $-0.000$ & $0.000$ & $-0.005$ & $-0.018$ \\
grboost & $0.000$ & $-0.001$ & $0.004$ & $-0.034$ & $-0.006$ & $0.000$ & $0.008$ & $0.073$ & $-0.006$ & $0.001$ & $0.002$ & $0.107$ \\
xgboost & $0.001$ & \cellcolor[HTML]{d4edda} $-0.083^{***}$ & $0.386$ & \cellcolor[HTML]{d4edda} $-2.521^{***}$ & $0.000$ & \cellcolor[HTML]{d4edda} $-0.068^{***}$ & \cellcolor[HTML]{d4edda} $-1.307^{***}$ & \cellcolor[HTML]{d4edda} $-1.725^{***}$ & $-0.000$ & \cellcolor[HTML]{f8d7da} $0.015^{***}$ & \cellcolor[HTML]{d4edda} $-1.690^{***}$ & \cellcolor[HTML]{f8d7da} $0.796^{***}$ \\
deepsurv & \cellcolor[HTML]{d4edda} $0.009^{*}$ & \cellcolor[HTML]{d4edda} $-0.020^{***}$ & \cellcolor[HTML]{d4edda} $-0.867^{**}$ & \cellcolor[HTML]{d4edda} $-0.311^{**}$ & \cellcolor[HTML]{d4edda} $0.011^{*}$ & \cellcolor[HTML]{d4edda} $-0.026^{***}$ & \cellcolor[HTML]{d4edda} $-4.474^{***}$ & \cellcolor[HTML]{d4edda} $-0.570^{***}$ & $0.001$ & $-0.007$ & \cellcolor[HTML]{d4edda} $-3.608^{***}$ & \cellcolor[HTML]{d4edda} $-0.259^{**}$ \\
deephit & $-0.001$ & \cellcolor[HTML]{d4edda} $-0.032^{***}$ & $-0.358$ & \cellcolor[HTML]{d4edda} $-0.751^{***}$ & $-0.007$ & \cellcolor[HTML]{d4edda} $-0.007^{*}$ & $-0.000$ & \cellcolor[HTML]{f8d7da} $0.352^{***}$ & $-0.006$ & \cellcolor[HTML]{f8d7da} $0.024^{***}$ & $0.366$ & \cellcolor[HTML]{f8d7da} $1.102^{***}$ \\
survtrace & $0.002$ & \cellcolor[HTML]{d4edda} $-0.068^{***}$ & $0.180$ & \cellcolor[HTML]{d4edda} $-1.948^{***}$ & $-0.003$ & \cellcolor[HTML]{d4edda} $-0.049^{***}$ & $-0.475$ & \cellcolor[HTML]{d4edda} $-1.092^{***}$ & $-0.004$ & \cellcolor[HTML]{f8d7da} $0.019^{***}$ & \cellcolor[HTML]{d4edda} $-0.650^{*}$ & \cellcolor[HTML]{f8d7da} $0.854^{***}$ \\
coxkan & $-0.006$ & $-0.005$ & \cellcolor[HTML]{d4edda} $-2.652^{***}$ & $-0.080$ & \cellcolor[HTML]{f8d7da} $-0.010^{*}$ & $-0.003$ & \cellcolor[HTML]{d4edda} $-2.651^{***}$ & $-0.010$ & $-0.004$ & $0.002$ & $0.002$ & $0.071$ \\
coxtime & $0.002$ & \cellcolor[HTML]{d4edda} $-0.020^{***}$ & \cellcolor[HTML]{d4edda} $-0.987^{**}$ & \cellcolor[HTML]{d4edda} $-0.548^{***}$ & \cellcolor[HTML]{d4edda} $0.011^{*}$ & \cellcolor[HTML]{d4edda} $-0.027^{***}$ & \cellcolor[HTML]{d4edda} $-3.984^{***}$ & \cellcolor[HTML]{d4edda} $-0.846^{***}$ & $0.009$ & $-0.007$ & \cellcolor[HTML]{d4edda} $-2.986^{***}$ & \cellcolor[HTML]{d4edda} $-0.299^{**}$ \\

\bottomrule
\end{tabular}%
}
\end{table}

This phenomenon is strikingly evident in complex methods. As shown in Tab. \ref{tab:selection_switch_effects}, when shifting the optimization metric from C-index to IBS, XGBoost exhibits a substantial median IBS improvement of -0.083 ($pd=1.0$) and SurvTRACE improves by -0.068 ($pd=1.0$). Crucially, this massive gain in probabilistic accuracy comes at no cost to discrimination, as their corresponding C-index shifts remain statistically negligible ($pd \approx 0.5-0.6$). In contrast, standard C-index selection does not adequately regularize flexible models, leading to highly uncalibrated predictions.

\textbf{\circnumA{A.3} Time-to-Event Accuracy Requires its Own Model Selection Approach.} 
The necessity of target-specific selection extends to absolute time-to-event accuracy (MAE-PO). Unlike $C^{td}$ and IBS, which are inherently bounded, MAE-PO is expressed in native time units and is therefore unbounded, meaning a single strongly mispredicted survival time can inflate the error substantially. Due to heavy outliers and noise, MAE-PO variance is harder to explain (42.7\% residual variance), but the choice of selection criterion remains highly consequential. If model selection relies on C-index or IBS, whether a model's MAE-PO improves over the Kaplan-Meier (KM) baseline \citep{Kaplan1958NonparametricEstimation} is essentially a coin flip (50\%) (see Tab. \ref{tab:mae_rel_summary_all}). Selecting models explicitly to optimize MAE-PO raises this success rate (i.e., rate of outperforming the KM baseline) to 72\% while simultaneously curbing severe prediction outliers (see Tab. \ref{tab:mae_rel_summary_all}). Furthermore, Tab. \ref{tab:selection_switch_effects} demonstrates that transitioning from C-index to MAE-PO selection drastically reduces temporal errors for DeepSurv by 99\% and CoxTime by 98\% (calculated from the log-multiplier scale). We conclude that researchers must move beyond C-index-only optimization to accurately assess a modern architecture's full capabilities beyond ranking.

Once the evaluation objective and corresponding model selection criterion have been established, the remaining question is which methods are expected to perform best under the characteristics of the target dataset. We therefore turn to regime-conditioned analyses that explain how benchmark conclusions change across different dataset regimes. Practitioners can use these findings to narrow down a suitable method for their specific setup.

\textbf{\circnumB{B} Contextualizing Performance: The Role of Dataset Regimes}

In practice, a model's performance is deeply tied to dataset properties, including scale, dimensionality, and censoring rates. Through our Bayesian GLMM, we isolate these meta-features to show how model performance interacts with distinct data regimes. We highlight the danger of relying on aggregate summaries by visualizing the severe rank shifts that occur between small and large datasets. Furthermore, we expose a structural vulnerability in absolute metric evaluation, demonstrating that heavy censoring renders X-CAL scores artificially conservative.

\begin{wrapfigure}{r}{0.47\textwidth}
    \vspace{-12pt}
    \centering
    \includegraphics[width=\linewidth]{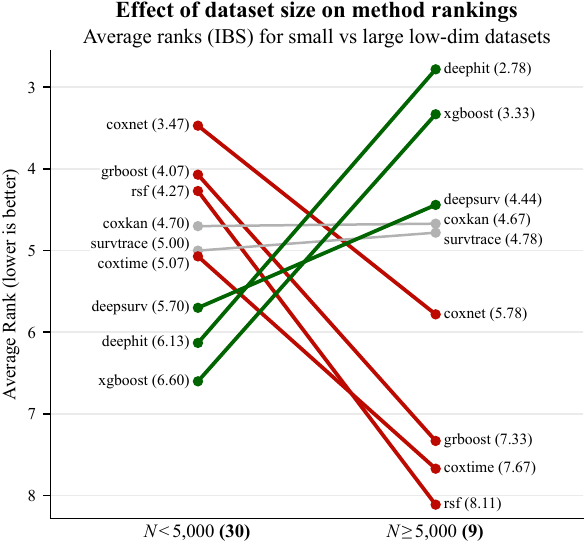}
    \caption{Average ranks (lower is better) for small ($N < 5k$) vs large ($N \geq 5k$) datasets.}
    \label{fig:rank_shifts_IBS_lowdim_small_vs_large}
    \vspace{-14pt}
\end{wrapfigure}

\textbf{\circnumB{B.1} Dataset Scale and Dimensionality Govern Method Performance.} 
The structural properties of a dataset fundamentally dictate which architectural family will succeed. Regarding sample scale ($\log N$), highly flexible architectures exhibit strong, consistent positive scaling for discriminative performance (Fig. \ref{fig:log_n_pd}). As sample sizes increase, models like XGBoost,  DeepHit, and SurvTRACE show substantial IBS improvements with median posterior effects ranging from +0.041 to +0.049 (all $pd \geq 0.99$). While all models seem to benefit from larger sample sizes, we have to emphasize that their effect strengths are vastly different. This scaling effect is so severe that it entirely upends model rankings: as demonstrated in Fig. \ref{fig:rank_shifts_IBS_lowdim_small_vs_large}, DeepHit leaps from near last place on small datasets to first place on large-scale datasets. Please note that methods that did fail on large datasets were automatically assigned the worst rank for that specific dataset.

\setlength{\columnsep}{10pt}

\begin{figure}[H]
    \centering
    \includegraphics[width=\linewidth]{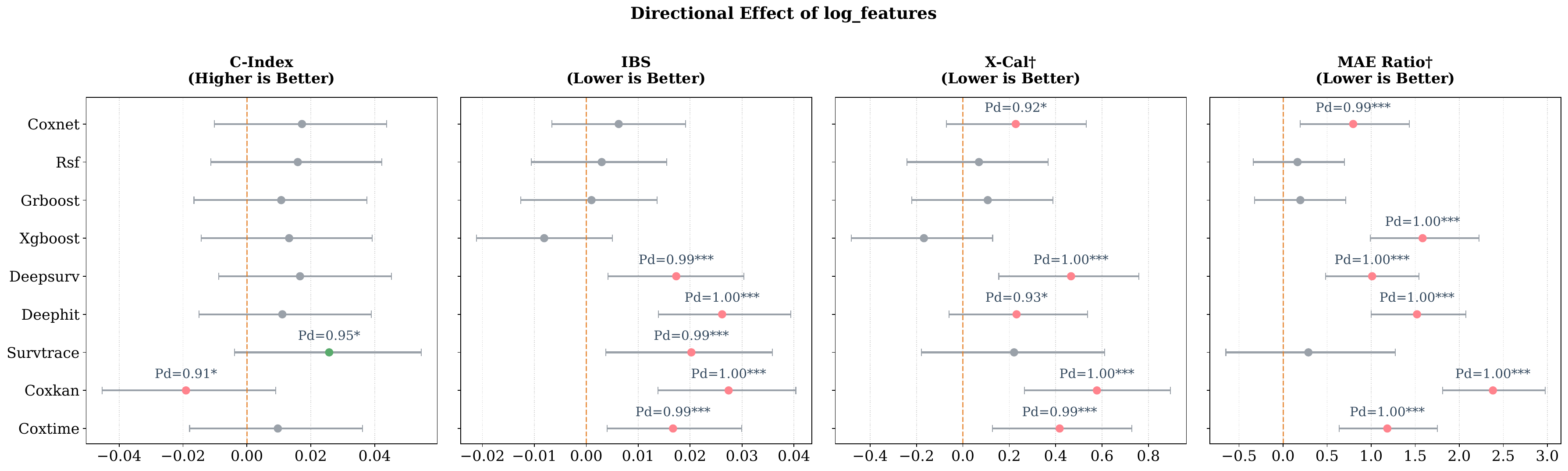}
    
    \caption{Effects of changing the Log-Feature size by one standard deviation for each method on target metrics. Significant ($pd$ > 0.9) results are colored. Red: detrimental effect, green: beneficial effect. A dagger ($\dagger$) denotes metrics that are reported on a log-multiplier scale.}
    \label{fig:log_features_pd}
\end{figure}

However, this architectural advantage reverses in high-dimensional feature spaces ($\log P$). Our posterior projections reveal a severe vulnerability of deep learning models to high dimensionality concerning probabilistic accuracy (Fig. \ref{fig:log_features_pd}). DeepHit (+0.026), CoxKAN (+0.027), and SurvTRACE (+0.02) all exhibit significant IBS deterioration with certainty ($pd \geq 0.99$).  This suggests that while deep survival models effectively leverage large sample sizes to improve ranking, they lack the regularization required to control their predictive distributions for high-dimensional settings. We further investigate the effect of regularization through an additional dropout ablation study on a representative high-dimensional dataset, presented Appendix \ref{app:reg_ablation}, Table \ref{tab:regularization_ablation}, where the ablation study indicates that regularization in general can be beneficial.

\textbf{\circnumB{B.2} The Calibration Fallacy: Relative vs Absolute Utility.} 
Our analysis exposes a critical structural vulnerability in standard survival evaluation practices: raw absolute calibration metrics are heavily biased by the underlying censoring regime. Our Bayesian GLMM reveals a systematic negative effect of the censoring rate on X-CAL (Fig. \ref{fig:censoring_rate_pd}). As censoring increases, a model's apparent calibration artificially improves.  This phenomenon can be explained by a mathematical ``smearing effect'': survival probabilities are distributed over unobserved time horizons, artificially smoothening the distribution towards good calibration.

\setlength{\columnsep}{15pt}
\setlength{\intextsep}{0pt} 
\begin{wrapfigure}{r}{0.45\textwidth} %
    \centering
    \captionsetup{font=small, labelfont=bf, justification=centering} 
    \captionof{table}{D-Calibration rejection rates at low versus high censoring levels.} %
    \label{tab:dcal_censoring}
    \vspace{-5pt}
    \resizebox{0.8\linewidth}{!}{%
        \begin{tabular}{lcc}
            \toprule
            & \multicolumn{2}{c}{\textbf{Rejection Rate}} \\
            \cmidrule(lr){2-3}
            \textbf{Method} & $\le 40\%$ Cens. & $\ge 80\%$ Cens. \\
            \midrule
            CoxNet   & 24.6\% & 0.0\%  \\
            DeepSurv & 27.7\% & 0.0\%  \\
            DeepHit  & 33.8\% & 13.3\% \\
            \bottomrule
        \end{tabular}%
    }
\end{wrapfigure}

While X-CAL remains a robust and highly informative metric for the \textit{relative} ranking of models on the exact same dataset, its absolute value becomes unreliable under heavy censoring. Consequently, standard D-Calibration tests become overly conservative and susceptible to false negatives.  As shown in Tab. \ref{tab:dcal_censoring}, the rejection rate for miscalibration in DeepHit drops from 33.8\% on low-censoring datasets ($\le 40\%$) to just 13.3\% on high-censoring datasets ($\ge 80\%$). Practitioners must be wary: a model is not necessarily well-calibrated simply because a test fails to reject it on a highly censored dataset. This finding matches the identified flaws of D-Calibration in a recently published paper by \citet{simonsen2025calibration}.

\section{Conclusion}
\label{sec:conclusion}
The prevailing consensus that classical Cox models remain competitive is an artifact of narrow evaluation conventions, not an absolute methodological truth. By benchmarking 10 methods across 49 diverse datasets under varying model selection criteria, we demonstrate that survival model performance is fundamentally regime-dependent and that the selection metric strongly shapes model behavior. 

\textbf{Beyond Single-Number Summaries.} We challenge the convention of summarizing survival method performance reporting through global average summaries, particularly when models are uniformly optimized on the C-index but evaluated across distinct metrics. While average summary statistics provide a convenient overview (we provide standard heatmaps and average ranks in Appendix \ref{app:general_suppl_material} for reference), they are inherently shaped by the structural composition of the benchmark, including the distribution of dataset scales, dimensionalities, and censoring rates. As our Bayesian GLMMs show, highly flexible methods benefit strongly from IBS-based selection, a dynamic completely hidden under standard C-index optimization. Rankings also shift dramatically with dataset scale: DeepHit, for instance, moves from second last place on average-sized datasets to first on large-scale ones when using IBS as the model selection criterion (see Fig. \ref{fig:rank_shifts_IBS_lowdim_small_vs_large}). Global rankings should therefore be treated as a description of the benchmark corpus, not as intrinsic properties of the algorithms. We advocate for \textbf{regime-conditioned reporting}: stratifying performance by dataset characteristics and evaluation target, and treating model selection criteria as an independent experimental variable.

\textbf{Practical Recommendations.} Based on our findings, we offer the following guidance for practitioners: CoxNet remains a robust default choice, especially on small to mid-sized datasets, and offers interpretability. On large datasets, DeepHit, DeepSurv, and XGBoost consistently outperform Linear Cox models \circnumB{B.1}. When accurate event-time prediction matters, MAE-PO-based model selection is essential \circnumA{A.3}; SurvTRACE performs particularly well under this regime, possibly due to its auxiliary ``length-of-stay'' objective (essentially a squared error term for predicted survival times). RSF provides the strongest trade-off between discrimination, calibration, and robustness, with stable MAE-PO and no observed outliers, but it does not scale well to datasets with very large sample size~$N$. In high-dimensional regimes, RSF, CoxNet, and GrBoost remain reliable, while DL methods often degrade or even fail due to compute constraints \circnumB{B.1}. IBS-based selection consistently improves calibration \circnumA{A.2}. Under high censoring, practitioners should be aware of the \textit{Calibration Fallacy} \circnumB{B.2}. For interpretable yet flexible modeling, CoxKAN provides symbolic log-hazard expressions, while being a competitive and stable neural method, especially for ranking assessment.

\textbf{Limitations.} The study is subject to several limitations. The benchmark, despite containing 17 more datasets than the largest previous benchmark, particularly expanding in high-dimensional and large-$N$ regimes, remains overall skewed toward small and mid-sized datasets. A uniform CPU-only environment was fixed for all methods, and a fixed per-fold compute budget was enforced due to the model selection metrics being CPU-bound. Some aspects are especially detrimental to deep learning methods. Hyperparameter grids were informed by published papers, reference implementations, and prior benchmarks to the best of our knowledge, but may nonetheless miss optimal configurations and systematically underestimate individual methods' potential. Lastly, the benchmark is restricted to right-censored, single-event, static tabular data.

\textbf{Outlook.} Our findings suggest that time-to-event prediction needs to be rethought end-to-end, not only in terms of which metrics are reported, but also how model selection is performed and how metric implementations are engineered to make proper scoring rules practical at scale. Currently, probabilistic accuracy and time-to-event accuracy still diverge under optimization: improving one can degrade the other, forcing target-specific selection rather than yielding jointly accurate predictions. Closing this gap, whether through joint losses, unified metrics, or architectures that structurally couple calibration with temporal precision, remains an open challenge. We showed that aligning the selection criterion with the target application can unlock larger gains than switching to a more complex architecture. The importance of these findings will only grow as foundation models enter survival analysis and datasets scale by orders of magnitude.
\newpage

\bibliographystyle{plainnat}
\bibliography{bibliography}

\clearpage
\appendix

\section{Related Work}
\label{app:related_work}
\begin{table}[htpb]
\centering
\caption{Summary of existing survival analysis benchmarking studies.}
\label{tab:benchmark_summary}
\resizebox{\linewidth}{!}{%
\begin{tabular}{l c c c c c c c c c c c}
\toprule

\textbf{Study} & 
\makecell{\textbf{\#Meth.} \\ \textcolor[HTML]{BD0A00}{\textbf{\#DL}}} & 
\textbf{\#Data} & 
\textbf{$N$} & 
\textbf{$P$} & 
\makecell{\textbf{HP}\\\textbf{Tuning}} & 
\makecell{\textbf{Selection}\\\textbf{Criteria}} & 
\textbf{$C^{td}$} & 
\makecell{\textbf{Prob.}\\\textbf{Acc}} & 
\makecell{\textbf{Time}\\\textbf{Acc}} & 
\textbf{Cal} &
\textbf{Impl.} \\
\midrule

Herrmann et al. (2020) & \makecell{13 \\ \textcolor[HTML]{BD0A00}{0}} & 18 & (35, 735) & (90154, 103369) & Yes & \makecell{Internal \\ Objective} & No & Yes & No & No & R \\
Zhang et al. (2022) & \makecell{22 \\ \textcolor[HTML]{BD0A00}{3}} & 16 & (41, 3323) & (7, 19818) & No & - & No & Yes & No & No & R \\
Abubakari et al. (2024) & \makecell{9 \\ \textcolor[HTML]{BD0A00}{1}} & 4 & (58, 228) & (8, 19818) & No & - & No & No & No & No & R \\
Chien et al. (2024) & \makecell{7 \\ \textcolor[HTML]{BD0A00}{3}} & 6 & (985, 10536) & (25, 100) & No & - & No & No & No & No & - \\
B\'askay et al. (2025) & \makecell{5 \\ \textcolor[HTML]{BD0A00}{1}} & 4 & (137, 9105) & (6, 40) & Yes & \makecell{Modified\\C-index} & No & No & No & No & Python \\
Birolo et al. (2025) & \makecell{8 \\ \textcolor[HTML]{BD0A00}{3}} & 6 & (1903, 6524) & (8, 9) & Yes & \makecell{$C^{td}$} & Yes & Yes & No & No & Python \\
Green et al. (2025) & \makecell{5 \\ \textcolor[HTML]{BD0A00}{2}} & 16 & (208939, 644296) & 128 & Yes & \makecell{$C^{td}$} & \makecell{only DL-HP \\ selection} & No & No & No & R \\
Burk et al. (2026) & \makecell{19 \\ \textcolor[HTML]{BD0A00}{0}} & 32 & (137, 52422) & (2, 26) & Yes & C-index, IBS & No & Yes & No & Yes & R \\
\textbf{\textcolor[HTML]{3070B3}{Ours (2026)}} & 
\textbf{\makecell{10 \\ \textcolor[HTML]{BD0A00}{5}}} & 
\textbf{49} & 
\textbf{(26, 100000)} & 
\textbf{(3, 15500)} & 
\textbf{Yes} & 
\textbf{\makecell{\boldmath $C^{td}$, IBS,\\ MAE-PO}} &
\textbf{Yes} & 
\textbf{Yes} & 
\textbf{Yes} & 
\textbf{Yes} & 
\textbf{Python} \\

\bottomrule
\end{tabular}%
}
\end{table}

\clearpage
\section{Datasets}
\label{app:datasets}
\begin{table*}[htbp]
\centering
\scriptsize
\caption{Overview of all 49 datasets included in the benchmark. $N$ denotes the number of observations, $P$ the number of features after preprocessing, and Cens.\ the percentage of right-censored observations. Tier~2 datasets are marked with $\dagger$.}
\label{tab:dataset_overview}
\begin{tabular}{llrrrl}
\toprule
Dataset & Source & $N$ & $P$ & Cens.\ (\%) & Domain \\
\midrule
\texttt{acath}              & SurvSet      &   2{,}230 &      3 & 34 & Clinical \\
\texttt{ADNI}$^\dagger$     & ADNI         &       760 &     29 & 63 & Clinical \\
\texttt{Aids2}              & SurvSet      &   2{,}810 &     11 & 38 & Clinical \\
\texttt{AML\_Bull}          & SurvSet      &       115 &  6{,}266 & 43 & Omics \\
\texttt{borrow\_repay}$^\dagger$ & FinSurvival &  54{,}147 &     70 & 40 & Financial \\
\texttt{breast}             & SurvSet      &       100 &      4 & 74 & Clinical \\
\texttt{burn}               & SurvSet      &       154 &     13 & 69 & Clinical \\
\texttt{cancer}             & SurvSet      &       169 &     18 & 28 & Clinical \\
\texttt{cgd}                & SurvSet      &       128 &     17 & 66 & Clinical \\
\texttt{chop}               & SurvSet      &       412 &  3{,}833 & 60 & Omics \\
\texttt{cost}               & SurvSet      &       518 &     13 & 22 & Clinical \\
\texttt{dataDIVAT1}         & SurvSet      &   5{,}943 &     14 & 84 & Clinical \\
\texttt{dataDIVAT3}         & SurvSet      &   4{,}267 &     16 & 94 & Clinical \\
\texttt{DBCD}               & SurvSet      &       295 &  4{,}912 & 73 & Omics \\
\texttt{Dialysis}           & SurvSet      &   6{,}805 &     16 & 76 & Clinical \\
\texttt{DLBCL}              & SurvSet      &       240 &  7{,}343 & 43 & Omics \\
\texttt{divorce}            & SurvSet      &   3{,}371 &      4 & 69 & Socioeconomic \\
\texttt{e1684}              & SurvSet      &       284 &      3 & 31 & Clinical \\
\texttt{flchain}            & SurvSet      &   6{,}521 &     25 & 70 & Clinical \\
\texttt{Framingham}         & SurvSet      &   4{,}658 &     16 & 69 & Clinical \\
\texttt{gse1992}            & SurvSet      &       123 & 15{,}500 & 72 & Omics \\
\texttt{gse3143}            & SurvSet      &       158 &  8{,}491 & 68 & Omics \\
\texttt{gse4335}            & SurvSet      &       113 & 12{,}769 & 66 & Omics \\
\texttt{hdfail}             & SurvSet      &  52{,}422 &     15 & 95 & Engineering \\
\texttt{hepatoCellular}     & SurvSet      &       101 &     42 & 55 & Clinical \\
\texttt{kkbox\_100k}$^\dagger$ & pycox     & 100{,}000 &     27 & 35 & Commercial \\
\texttt{LeukSurv}           & SurvSet      &   1{,}043 &     16 & 16 & Clinical \\
\texttt{MCLcleaned}         & SurvSet      &        92 &    574 & 30 & Omics \\
\texttt{metabric}$^\dagger$ & cBioPortal   &   1{,}979 &     13 & 42 & Clinical \\
\texttt{micro.censure}      & SurvSet      &       117 &     80 & 78 & Omics \\
\texttt{mimic\_iv}$^\dagger$& PhysioNet    &  54{,}724 &     66 & 91 & Clinical \\
\texttt{nki70}              & SurvSet      &       144 &     76 & 67 & Omics \\
\texttt{NSBCD}              & SurvSet      &       115 &    549 & 67 & Omics \\
\texttt{nwtco}              & SurvSet      &   4{,}028 &      9 & 86 & Clinical \\
\texttt{ovarian}            & SurvSet      &        26 &      4 & 54 & Clinical \\
\texttt{pbc}                & SurvSet      &       312 &      6 & 60 & Clinical \\
\texttt{pharmacoSmoking}    & SurvSet      &       113 &     15 & 32 & Socioeconomic \\
\texttt{prostateSurvival}   & SurvSet      &  14{,}065 &      6 & 94 & Clinical \\
\texttt{rhc}                & SurvSet      &       634 &     64 & 58 & Clinical \\
\texttt{scania}             & SurvSet      &   1{,}931 &      8 & 44 & Engineering \\
\texttt{smarto}             & SurvSet      &   1{,}021 &     31 & 89 & Omics \\
\texttt{support2}$^\dagger$ & SUPPORT      &   8{,}873 &     19 & 32 & Clinical \\
\texttt{UnempDur}           & SurvSet      &   3{,}241 &      6 & 39 & Socioeconomic \\
\texttt{Unemployment}       & SurvSet      &       450 &      7 & 44 & Socioeconomic \\
\texttt{vdv}                & SurvSet      &        78 &  4{,}672 & 56 & Omics \\
\texttt{veteran}            & SurvSet      &       137 &      8 &  7 & Clinical \\
\texttt{vlbw}               & SurvSet      &       257 &     30 & 86 & Clinical \\
\texttt{Z243}               & SurvSet      &       100 &     15 &  0 & Clinical \\
\texttt{zinc}               & SurvSet      &       431 &     15 & 81 & Clinical \\
\bottomrule
\end{tabular}
\end{table*}

\paragraph{Preprocessing Details - Tier 2 Datasets.} 
\label{app:dataset_preprocessing}
Tier 2 datasets undergo dataset-specific preprocessing before entering the shared pipeline. For METABRIC, this includes binary encoding of receptor status, grade normalization, and median imputation of missing clinical values. For KKBox, missing gender and city values are filled with placeholder categories, and remaining numeric missingness is median-imputed. SUPPORT2 follows the established preprocessing from \citet{Katzman2018DeepSurv} which manually selects 14 features and applies its own scaling and encoding - it bypasses the shared pipeline but still passes through the NZV and high-correlation filters. MIMIC IV preprocessing follows \citet{tang2021sodenscalablecontinuoustimesurvival}, extracting 26 features from the first 24 hours of ICU data. In all cases, the dataset-specific steps produce a dataframe with \texttt{num\_}/\texttt{fac\_} prefixed columns, which then enters the shared pipeline.

\paragraph{ADNI.} The \textit{Alzheimer's Disease Neuroimaging Initiative} (ADNI) is a longitudinal, multi-site study launched in 2003 with the objective of determining whether serial MRI, PET, CSF biomarkers, and clinical assessments can be combined to measure the progression of mild cognitive impairment (MCI) and early Alzheimer's disease \citep{Petersen2010ADNI}. The raw data requires an application process \footnote{\url{https://adni.loni.usc.edu/data-samples/adni-data/}} but can be obtained for free after passing it. Brain volumetric features were obtained through \textit{FreeSurfer} \citep{Fischl2012FreeSurfer}. The event of interest is conversion from MCI to Alzheimer's disease, with a censoring rate of 62.9\%. The mean age is 73.5 years, 60.4\% are male, and the median follow-up time is 2 years.

\paragraph{Borrow-Repay.} To include a financial survival analysis setting, we include a dataset from the \textit{FinSurvival} suite by \citet{green2026finsurvivalsuitelargescale}, which models among other things lending and borrowing events in decentralized finance. The original FinSurvival benchmark uses predefined train/test splits that are incompatible with our cross-validation protocol and carries a risk of entity-level leakage (multiple observations per user). We therefore merge the original splits into a single dataset, restrict to the first transaction per user, and drop user-history features that become degenerate under this restriction and that are uninformative at a user's first interaction (e.g., time since first transaction). Time is converted from seconds to days. This substantially changes the dataset characteristics: the censoring rate shifts from approximately 14\% to approximately 40\%, and the task shifts from predicting repayment from behavioral history to predicting it from loan characteristics (amount or interest rate), market-level activity, and temporal context (e.g., time of the day). 

\paragraph{KKBox.} The KKBox dataset models customer churn for the KKBox music streaming service, where the event of interest is subscription cancellation. It originates from the \textit{WSDM KKBox Churn Prediction Challenge} and was adapted for survival analysis by Kvamme et al.\ \citep{kvamme2019timetoeventpredictionneuralnetworks, kvamme2019brierscoreadministrativecensoring}. The data is accessed via \texttt{pycox.datasets.kkbox}

With $N = 2{,}814{,}735$ subjects and $P = 27$ features, KKBox is by far the largest dataset in the benchmark; a subsampled version with $N = 100{,}000$ is used for the standardized evaluation, following \citet{alberge2025survival}. Before the shared pipeline, missing categorical values are filled with placeholder categories (gender $\to$ \texttt{"n/a"}, city and registration channel $\to$ \texttt{"0"}), and remaining numeric missingness is median-imputed.

\paragraph{METABRIC.} The \textit{Molecular Taxonomy of Breast Cancer International Consortium }(METABRIC) dataset was obtained from cBioPortal \footnote{\url{https://www.cbioportal.org/study/summary?id=brca\_metabric}} \citep{Curtis_2012_metabric}, \citep{Pereira_2016_metabric}. Clinical data and PAM50 gene expression measurements (50 genes; see below) were downloaded separately and merged on sample identifier. The cohort comprises 1{,}980 breast cancer patients after outcome filtering, with overall survival as the endpoint (event rate 57.8\%, median follow-up 116.5 months). The raw clinical features are age at diagnosis, tumor size, number of positive lymph nodes, neoplasm histologic grade, and receptor status for ER, PR, and HER2. Before the shared pipeline, receptor status variables are encoded as binary numerics, histologic grade is normalized to $[0, 0.5, 1]$ via $(g-1)/2$ and excluded from z-score scaling, and missing clinical values are median-imputed. Rows with missing gene expression are dropped.

The PAM50 gene signature consists of 50 genes commonly used for breast cancer subtype classification. The following list corresponds to the genes used in the METABRIC dataset variants.
\begin{multicols}{3}
\begin{itemize}
    \item UBE2T
    \item BIRC5
    \item NUF2
    \item CDC6
    \item CCNB1
    \item TYMS
    \item MYBL2
    \item CEP55
    \item MELK
    \item NDC80
    \item RRM2
    \item UBE2C
    \item CENPF
    \item PTTG1
    \item EXO1
    \item ORC6
    \item ANLN
    \item CCNE1
    \item CDC20
    \item MKI67
    \item KIF2C
    \item ACTR3B
    \item MYC
    \item EGFR
    \item KRT5
    \item PHGDH
    \item CDH3
    \item MIA
    \item KRT17
    \item FOXC1
    \item SFRP1
    \item KRT14
    \item ESR1
    \item SLC39A6
    \item BAG1
    \item MAPT
    \item PGR
    \item CXXC5
    \item MLPH
    \item BCL2
    \item MDM2
    \item NAT1
    \item FOXA1
    \item BLVRA
    \item MMP11
    \item GPR160
    \item FGFR4
    \item GRB7
    \item TMEM45B
    \item ERBB2
\end{itemize}
\end{multicols}

\paragraph{MIMIC IV.} The MIMIC dataset is derived from the publicly available \textit{Medical Information Mart for Intensive Care} database and provides anonymized clinical data of patients admitted to an intensive care unit (ICU) \citep{tang2021sodenscalablecontinuoustimesurvival}. Access requires completion of a CITI Program training course\footnote{\url{https://about.citiprogram.org/}}, after which the data can be obtained from PhysioNet\footnote{\url{https://physionet.org/content/mimiciv/3.1/}}. MIMIC-IV requires credentialed access, therefore we solely provide the preprocessing steps and no raw MIMIC data. Preprocessing follows \citet{tang2021sodenscalablecontinuoustimesurvival}, originally described for MIMIC-III, and \citet{PURUSHOTHAM2018112} for MIMIC-IV. The cohort comprises adult patients ($\geq$18) at their first ICU stay who are alive 24 hours after admission. The event of interest is in-hospital mortality, with the event time observed if a death record exists in the database and the censored time defined as the last hospital discharge time otherwise. Time is measured in hours from ICU admission, with administrative censoring applied at 30 days (720,h). 28 features are extracted from the first 24 hours of clinical data: 7 vital signs from chart events and 12 laboratory values - each mean-aggregated over the observation window - plus 8 demographic categoricals and age. Numerical features with $>$50\% missingness are dropped; missing categoricals are filled with a placeholder category; remaining numeric missingness is handled via listwise deletion. The resulting dataset contains $N = 54{,}724$ subjects, $P = 66$ encoded features, and a censoring rate of 91.2\%, providing a large-scale ICU mortality prediction setting.

\paragraph{SUPPORT2.} The SUPPORT2 dataset originates from the \textit{Study to Understand Prognoses Preferences Outcomes and Risks of Treatments} and models time to death for seriously ill hospitalized patients \citep{Knaus1995SUPPORT, Katzman2018DeepSurv}. Although a version of this dataset exists in SurvSet, we use the established preprocessing from \citet{Katzman2018DeepSurv} which is implemented in the survival analysis introduction repository by \citet{chen2024introductiondeepsurvivalanalysis}. It selects 14 clinically motivated features (age, sex, race, number of comorbidities, diabetes, dementia, cancer status and seven physiological measure). This preprocessing applies its own encoding and scaling: sex is binarized, cancer status mapped to an ordinal scale and normalized by its maximum, number of comorbidities divided by 9, eight continuous features z-scored, and race one-hot encoded without dropping a reference category. Rows with any missing values are dropped. After encoding, the dataset contains $N = 8{,}873$ subjects with $P = 19$ features and an event rate of 68.0\%. Time is measured in days.

\clearpage
\section{Methods, Implementations and Evaluations}
\label{app:methods_and_imp}
\subsection{Reproducibility}
\label{app:reproducibility}
All experiments were conducted using Python 3.10.19 on Linux x86\_64 (\texttt{glibc} 2.31) on a CPU-based machine (32 CPU Cores, 64GB RAM). Key dependencies include \texttt{PyTorch} 2.5.1, \texttt{NumPy} 1.26.4, \texttt{pandas} 2.3.3, \texttt{scikit-learn} 1.7.2, \texttt{pycox} 0.2.3, and \texttt{torchtuples} 0.2.2. 

\subsection{Method Variants and (Hyper)parametrization}
\label{app:hyperparametrization}

This section details the specific implementation and parameterization logic for the evaluated methods. Tab. \ref{tab:method_overview} links each method to its underlying Python library and model class, while Tab. \ref{tab:hp_grids} explicitly defines the hyperparameter search spaces ($\mathcal{G}$). 

For all deep learning architectures and XGBoost, checkpoint selection is performed dynamically via validation-based early stopping (patience $= 10$ epochs/rounds) to prevent overfitting.

Relevant libraries included in our experiments are \texttt{survhive} \citep{Birolo_2025_beyondCox}, \texttt{scikit-survival} \citep{Polsterl2020ScikitSurvival}, \texttt{xgboost} \citep{Chen_2016_XGboost} and \texttt{xgbse} \citep{xgbse2020github}, \texttt{pycox} \citep{kvamme2019timetoeventpredictionneuralnetworks}, \texttt{survtrace} \citep{survtrace_2022_wang} and \texttt{coxkan} \citep{Knottenbelt_2025}.

\begin{itemize}
    \item \textbf{CoxPH} \citep{Cox1972RegressionModels}: Unregularized, hyperparameter-free baseline accessed via \texttt{scikit-survival} and \texttt{survhive}.
    \item \textbf{CoxNet} \citep{Simon2011RegularizationPathsCox}: Elastic net regularized Cox via \texttt{scikit-survival} and \texttt{survhive}. The regularization trade-off (\texttt{l1\_ratio}) is tuned from Ridge to Lasso, utilizing a data-dependent penalty path for \texttt{alpha}.
    \item \textbf{RSF} \citep{Ishwaran2008RandomSurvivalForests} \textbf{\& GrBoost} \citep{Friedman2001GradientBoosting}: Random Survival Forest and Gradient Boosting via \texttt{scikit-survival} and \texttt{survhive}. Both tune structural tree-ensemble parameters (estimators, depth, split/leaf constraints).
    \item \textbf{XGBoost} \citep{Chen_2016_XGboost}: Native \texttt{xgboost} implementation with survival functions extracted via \texttt{xgbse}. Optimized under both Cox and Accelerated Failure Time (AFT) objectives. Survival functions are constructed via the Breslow estimator (Cox) or a closed-form distribution (AFT, detailed in Appendix \ref{app:xgboost}).
    \item \textbf{DeepSurv} \citep{Katzman2018DeepSurv}: Single-hidden-layer Cox MLP implemented via \texttt{pycox}.
    \item \textbf{DeepHit} \citep{Lee2018DeepHit}: Discrete-time network via \texttt{pycox}. The temporal discretization strategy (unique event times vs. 128 quantile-based bins) is treated as a structural hyperparameter, alongside the ranking loss weight \texttt{alpha}.
    \item \textbf{CoxTime} \citep{kvamme2019timetoeventpredictionneuralnetworks}: Time-dependent neural network via \texttt{pycox}, utilizing case-control sampling to dynamically approximate the partial likelihood.
    \item \textbf{SurvTRACE} \citep{survtrace_2022_wang}: Tabular transformer via \texttt{survtrace}. Treats numerical features as individual tokens mapped to multi-head self-attention and utilizes quantile-based time discretization.
    \item \textbf{CoxKAN} \citep{Knottenbelt_2025}: Kolmogorov-Arnold Network via \texttt{coxkan}. Replaces standard linear layers with learnable univariate activations on spline grids, tuned via the \texttt{grid} and \texttt{base\_fun} parameters.
\end{itemize}

\begin{table*}[htbp]
\centering
\scriptsize
\setlength{\tabcolsep}{4pt}
\caption{Overview of all methods included in the benchmark. \emph{Family} distinguishes linear Cox, tree-based, and DL. \emph{Library} indicates the Python package used for model training; \emph{Model Class} specifies the concrete class or API call. DeepHit is trained with two discretization variants (unique event times and 128 quantile-based bins). The libraries include \texttt{survhive} \citep{Birolo_2025_beyondCox}, \texttt{scikit-survival} \citep{Polsterl2020ScikitSurvival}, \texttt{xgboost} \citep{Chen_2016_XGboost} and \texttt{xgbse} \citep{xgbse2020github}, \texttt{pycox} \citep{kvamme2019timetoeventpredictionneuralnetworks}, \texttt{survtrace} \citep{survtrace_2022_wang} and \texttt{coxkan} \citep{Knottenbelt_2025}.}
\label{tab:method_overview}
\small
\begin{tabular}{@{} l l l l l @{}}
\toprule
Method & Family & Library & Model Class & Reference \\
\midrule
CoxPH & Cox Linear & survhive & \texttt{CoxPH} & scikit-survival \\
CoxNet & Cox Linear & survhive & \texttt{CoxNet} & scikit-survival \\
\addlinespace
RSF & Tree & survhive & \texttt{RSF} & scikit-survival \\
GradBoost & Tree & survhive & \texttt{GrBoostSA} & scikit-survival \\
XGBoost & Tree & xgboost + xgbse & \makecell[l]{\texttt{xgb.train} \\ (\texttt{survival:cox}, \texttt{survival:aft})} & xgboost \\
\addlinespace
DeepSurv & DL & pycox & \texttt{CoxPH} & pycox \\
DeepHit & DL & pycox & \texttt{DeepHitSingle} & pycox \\
SurvTRACE & DL & survtrace & \texttt{SurvTraceSingle} & survtrace \\
CoxKAN & DL & coxkan & \texttt{CoxKAN} & coxkan \\
CoxTime & DL & pycox & \texttt{CoxTime} & pycox \\
\bottomrule
\end{tabular}
\end{table*}

\subsection{Hyperparameter Search Spaces}
\label{app:hp_grids}
Hyperparameter grids were designed from published method papers and are stored in per-method YAML configuration files. Experiments were executed exclusively on CPUs without GPU acceleration. Depending on the method and dataset, the average runtime per hyperparameter configuration ranged from below one second for classical methods to several seconds for deep learning approaches, with an overall mean runtime of approximately 2 seconds per hyperparameter combination across all evaluated methods and datasets.

\begin{table*}[htbp]
\centering
\scriptsize
\setlength{\tabcolsep}{3.5pt}
\caption{Hyperparameter search spaces for all methods.}
\label{tab:hp_grids}
\small
\begin{tabular}{@{} l l l r @{}}
\toprule
Method & Hyperparameter & Search space & $|\mathcal{G}|$ \\
\midrule
CoxPH
    & \texttt{alpha}              & $\{0.0\}$                          & \multirow{2}{*}{1} \\
    & \texttt{ties}               & $\{\text{efron}\}$                 & \\
\midrule
CoxNet
    & \texttt{l1\_ratio}          & $\{0.05,\, 0.1,\, 0.25,\, 0.5,\, 0.75,\, 0.9,\, 0.99\}$ & \multirow{2}{*}{var.} \\
    & \texttt{alpha}              & data-dependent path                & \\
\midrule
RSF
    & \texttt{n\_estimators}      & $\{50,\, 100,\, 300\}$             & \multirow{4}{*}{48} \\
    & \texttt{max\_depth}         & $\{-\!\!-,\, 3,\, 6,\, 10\}$      & \\
    & \texttt{min\_samples\_split}& $\{0.01,\, 0.1\}$                  & \\
    & \texttt{min\_samples\_leaf} & $\{0.05,\, 0.2\}$                  & \\
\midrule
GrBoost
    & \texttt{n\_estimators}      & $\{50,\, 100,\, 300\}$             & \multirow{4}{*}{48} \\
    & \texttt{max\_depth}         & $\{-\!\!-,\, 3,\, 6,\, 10\}$      & \\
    & \texttt{min\_samples\_split}& $\{0.01,\, 0.1\}$                  & \\
    & \texttt{min\_samples\_leaf} & $\{0.05,\, 0.2\}$                  & \\
\midrule
XGBoost-Cox
    & \texttt{learning\_rate}     & $\{0.1,\, 0.3\}$                   & \multirow{3}{*}{12} \\
    & \texttt{max\_depth}         & $\{-\!\!-,\, 6,\, 10\}$            & \\
    & \texttt{num\_boost\_round}  & $\{500,\, 1000\}$                   & \\
\midrule
XGBoost-AFT
    & \texttt{aft\_loss\_distribution}       & $\{\text{normal},\, \text{extreme}\}$ & \multirow{5}{*}{72} \\
    & \texttt{aft\_loss\_distribution\_scale} & $\{0.5,\, 1.0,\, 2.0\}$             & \\
    & \texttt{learning\_rate}                & $\{0.1,\, 0.3\}$                     & \\
    & \texttt{max\_depth}                    & $\{-\!\!-,\, 6,\, 10\}$              & \\
    & \texttt{num\_boost\_round}             & $\{500,\, 1000\}$                     & \\
\midrule
DeepSurv
    & \texttt{hidden}             & $\{16,\, 128,\, 1024\}$            & \multirow{5}{*}{36} \\
    & \texttt{learning\_rate}     & $\{0.001,\, 0.01,\, 0.1\}$         & \\
    & \texttt{batch\_size}        & $\{256,\, 2048\}$                   & \\
    & \texttt{dropout}            & $\{0.0,\, 0.2\}$                    & \\
\midrule
DeepHit
    & \texttt{hidden}             & $\{16,\, 128,\, 1024\}$            & \multirow{6}{*}{72} \\
    & \texttt{learning\_rate}     & $\{0.001,\, 0.01,\, 0.1\}$         & \\
    & \texttt{batch\_size}        & $\{256,\, 2048\}$                   & \\
    & \texttt{dropout}            & $\{0.0,\, 0.2\}$                    & \\
    & \texttt{alpha}              & $\{0.2,\, 1.0\}$                    & \\
\midrule
SurvTRACE
    & \texttt{hidden\_factor}     & $\{4,\, 16\}$                      & \multirow{5}{*}{8} \\
    & \texttt{num\_attention\_heads} & $\{2,\, 4\}$                    & \\
    & \texttt{hidden\_dropout}    & $\{0.0,\, 0.2\}$                    & \\
\midrule
CoxKAN
    & \texttt{lr}                 & $\{0.001,\, 0.01,\, 0.05\}$        & \multirow{7}{*}{48} \\
    & \texttt{base\_fun}          & $\{\text{silu},\, \text{linear}\}$   & \\
    & \texttt{num\_hidden}        & $\{0,\, 1\}$                         & \\
    & \texttt{grid}               & $\{3,\, 5\}$                          & \\
    & \texttt{batch\_size}        & $\{256,\, 2048\}$                     & \\
\midrule
CoxTime
    & \texttt{hidden}             & $\{16,\, 128,\, 1024\}$            & \multirow{5}{*}{36} \\
    & \texttt{learning\_rate}     & $\{0.001,\, 0.01,\, 0.1\}$         & \\
    & \texttt{batch\_size}        & $\{256,\, 2048\}$                   & \\
    & \texttt{dropout}            & $\{0.0,\, 0.2\}$                    & \\
\bottomrule
\end{tabular}
\end{table*}

\newpage

\subsection{XGBoost}
\label{app:xgboost}
Neither XGBoost objective produces survival functions directly; we construct them from raw model outputs. \\
\textbf{Cox objective:} The model outputs a linear predictor $\hat{\eta}_i$. We estimate the cumulative baseline hazard $\hat{H}_0(t)$ via the Breslow estimator at each unique training event time and obtain individual survival curves. \\
\textbf{AFT objective:} The tree ensemble predicts $\hat{\mu} = \mathcal{T}(\mathbf{x})$, the expected log survival time, under the model $\ln(T) = \mathcal{T}(\mathbf{x}) + \sigma Z$. 
With $z = \frac{\ln t - \hat{\mu}}{\sigma}$, the following formula shows how to derive $S(t)$. Starting from $S(t) = P(T > t)$ and substituting the AFT model:
\begin{equation}
    S(t) = P(T > t) = P\!\left(Z > \frac{\ln t - \hat{\mu}}{\sigma}\right) = 1 - F_Z(z)
\end{equation}
where $F_Z$ is the CDF of the chosen error distribution. This yields the following closed-form survival functions per distribution after plugging in the respective CDF:
\begin{align}
    S_{\text{normal}}(t)   &= \frac{1}{2}\left(1 - \mathrm{erf}\!\left(\frac{z}{\sqrt{2}}\right)\right), \\
    S_{\text{logistic}}(t) &= \frac{1}{1 + e^{-z}}, \\
    S_{\text{extreme}}(t)  &= \exp\!\left(-e^{z}\right),
\end{align}
where the extreme distribution corresponds to the standard Gumbel (minimum) distribution. The scale parameter $\sigma$ the distribution choice, and their corresponding PDFs (Tab. \ref{tab:aft-loss}) are all treated as hyperparameters in the grid search.
\begin{table}[h]
\centering
\scriptsize
\setlength{\tabcolsep}{4pt}
\caption[AFT loss distributions and their probability density functions directly taken from the \textit{xgboost for Survival Analysis}-Documentation]{AFT loss distributions and their probability density functions directly taken from the \textit{xgboost for Survival Analysis}-Documentation.\footnotemark}
\label{tab:aft-loss}
\begin{tabular}{@{} l l @{}}
\toprule
AFT loss distribution & Probability Density Function \\
\midrule
Normal   & $\frac{1}{\sqrt{2\pi}} \exp(-z^2/2)$ \\
Logistic & $\frac{e^{z}}{(1 + e^{z})^2}$ \\
Extreme  & $e^{z}\, e^{-e^{z}}$ \\
\bottomrule
\end{tabular}
\end{table}
\footnotetext{\url{https://xgboost.readthedocs.io/en/latest/tutorials/aft_survival_analysis.html}}

\newpage

\subsection{Evaluation}
\label{app:eval_time_grid}
\paragraph{Time grid construction.} The evaluation time grid is constructed from all unique training follow-up times (events and censored observations), trimmed to the 1st and 99th percentile of training times to avoid boundary instability of the IPCW weights, and constrained to the intersection of training and test time ranges. If the grid exceeds 500 points, it is subsampled to 500 via uniform spacing across unique times. All native survival curves -- which differ substantially across method families (CoxPH: training event times; DeepHit: discrete bins; SurvTRACE: quantile cuts) -- are mapped to this shared grid via linear interpolation with anchors at ($t = 0$, $S=1.0$) and ($10 \cdot t_{last}, S = 0.0$), following the implementation recommendations of \citet{Archetti2024InterpolationSurvival}.

Two baselines are computed per fold: a random baseline $S(t) = 0.5$ for all patients and times) and a Kaplan-Meier baseline fitted on the training data. Both are evaluated on the same grid and serve as a reference points for $IBS\_rel$ and $MAE\_rel$. 

\clearpage
\section{Analysis Methods}
\subsection{Bayesian GLMMs}
\subsubsection{Bayesian Generalized Mixed Effects Model (GLMM)}
\label{app:GLMM}

To rigorously isolate the effects of algorithmic choice, model selection strategy, and dataset characteristics on predictive performance, we employed a Bayesian mixed-effects modeling framework. This approach estimates the marginal effects of algorithms and optimization strategies while explicitly accounting for the varying baseline difficulties across our evaluation corpus.

Based on the geometric constraints of our evaluation metrics, we utilize the following likelihoods and link functions:

\begin{enumerate}
    \item \textbf{Discrimination and Brier Scores (Gaussian):} For the C-index and IBS, we utilized a Gaussian likelihood with an identity link function. While these metrics are bounded on $[0, 1]$, rigorous posterior predictive checks (\ref{app:posterior_analysis}) demonstrated that a Gaussian model provides an excellent approximation with minor boundary violations. The model is formulated as $Y_{ijk} \sim \mathcal{N}(\mu_{ijk}, \sigma_\epsilon^2)$, where $\mu_{ijk} = \eta_{ijk}$. This preserves the direct, additive interpretability of the marginal effects.
    
    \item \textbf{Calibration and Absolute Error (Gamma):} The X-CAL and MAE-PO metrics are strictly positive ($Y_{ijk} > 0$) and heavily right-skewed. To ensure stable inference and capture this geometry, we modeled these targets using a Gamma distribution parameterized by its mean $\mu_{ijk}$ and a shape parameter $\nu$, paired with a log link function. The generative model is $Y_{ijk} \sim \text{Gamma}(\mu_{ijk}, \nu)$, where $\log(\mu_{ijk}) = \eta_{ijk}$. Consequently, additive covariate effects in $\eta_{ijk}$ translate to multiplicative scaling effects on the expected error.
    
\end{enumerate}

All Bayesian models were fitted using Hamiltonian Monte Carlo via the \texttt{bambi} \citep{Capretto2022} and \texttt{PyMC} \citep{pymc2023} libraries in Python. Used versions: pymc 5.28.4, bambi 0.17.2.

\subsubsection{Probability of Direction}
\label{app:probability_of_direction}

The probability of direction ($pd$) quantifies posterior certainty about the sign of an effect or contrast $\Delta$ \citep{makowski2019indices}:

\begin{equation}
    pd(\Delta) = \max\left\{
        P(\Delta > 0 \mid \mathcal{D}),
        P(\Delta < 0 \mid \mathcal{D})
    \right\},
\end{equation}

where $\mathcal{D}$ denotes the observed data. We estimate these probabilities from the proportions of posterior draws on either side of zero. Values near $0.5$ indicate uncertainty about the direction, whereas values near $1$ indicate high directional certainty, independent of effect magnitude.

\subsubsection{Variance Decomposition}
\label{app:variance_decomposition}
For the variance partitioning shown in Fig. \ref{fig:combined_variance_partitioning}, we followed a Nakagawa-style decomposition \citep{nakagawa2013general,nakagawa2017coefficient}. Therefore, we partition the variance of each fitted GLMM into fixed-effect, dataset-level random-effect, and residual components (matching the parametrization of our GLMM). For each posterior draw, fixed-effect variance is calculated as the empirical variance of the fixed-effects linear predictor across benchmark observations. Random-effect variance is given by the estimated dataset random-intercept variance, while residual variance is obtained from the Gaussian error variance or, for Gamma models, the trigamma function of the shape parameter. Furthermore, for each posterior draw, we normalize the three variance components by their sum. Fig. \ref{fig:combined_variance_partitioning} reports posterior mean proportions and 94\% highest-density intervals (HDIs), evaluated on the original scale for Gaussian models and the log-link scale for Gamma models.

\subsubsection{Posterior Analysis}
\label{app:posterior_analysis}

The consensus in data analysis is that a simpler model should always be preferred. Yet simple can be defined along many axes. For an analysis paper in the field of benchmarking survival analysis methods, we argue that a Gaussian model should be prepared if it adequately (enough) describes the posterior distributions of the modeled variables. This is simply due to the fact, that the estimated coefficients are very easy to understand as they live in data space / scale and are additive.

That being said, in the following, we analyze the posterior probability curves of predicted models together with the actual data curves.

\begin{figure}[h]
    \vspace{1mm}
    \centering
    \includegraphics[width=\linewidth]{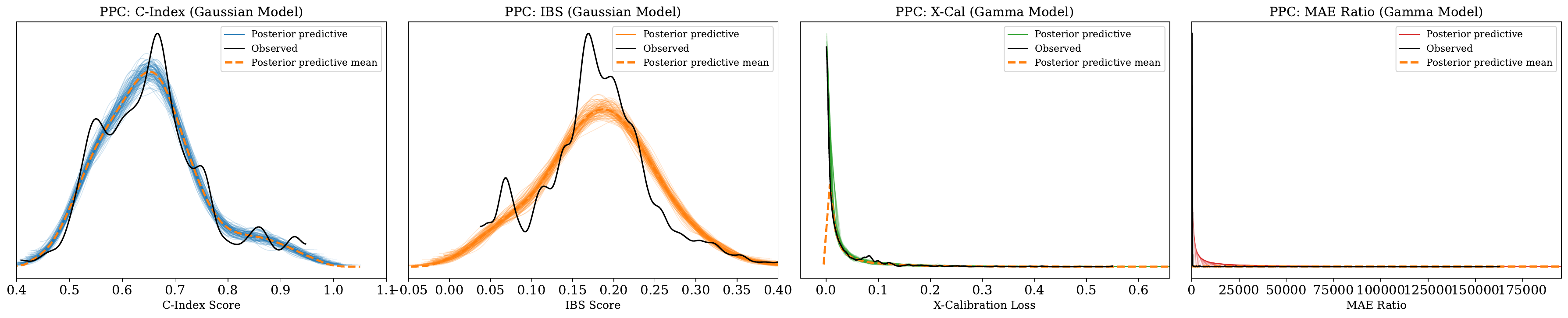}
    \caption{Posterior Predictive Check Used Assumptions in Main Body of Paper}
    \label{fig:posterior_predictive_all}
\end{figure}

\begin{figure}[h]
    \centering
    \vspace{2mm}
    \includegraphics[width=\linewidth]{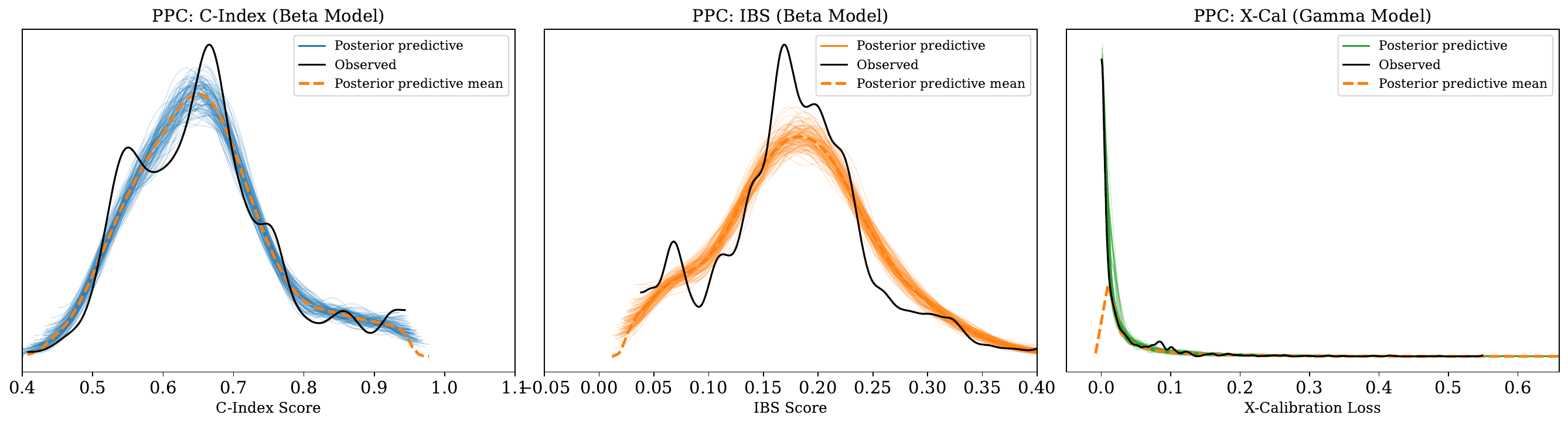}
    \caption{Posterior Predictive Check Beta for C-index and IBS}
    \label{fig:posterior_predictive_beta_gamma}
\end{figure}

For C-index and IBS, the correct modeling choices would be a Beta distribution, as both metrics live on limited intervals. But for these two metrics we see in Fig. \ref{fig:posterior_predictive_all} and \ref{fig:posterior_predictive_beta_gamma} that the Gaussian curves actually are very similar in total shape to their Beta variants. While they slightly transgress out of the valid ranges, we argue that this is still in an acceptable range. We especially confirmed that none of the findings and trends change when using Beta models for these two metrics in additional Bayesian pd and variance decomposition analyses.

For MAE-PO and X-CAL, however, we directly used Gamma distributions, as it was immediately apparent that simple Gaussian's are highly inaccurate given the data distributions. In these cases, we therefore accept the conceptually harder to grasp and explain log-multiplier effects in order to better respect their true distributions.


\subsection{Probability Direction Analysis}

\begin{figure}[htbp]
    \centering
    \includegraphics[width=\linewidth]{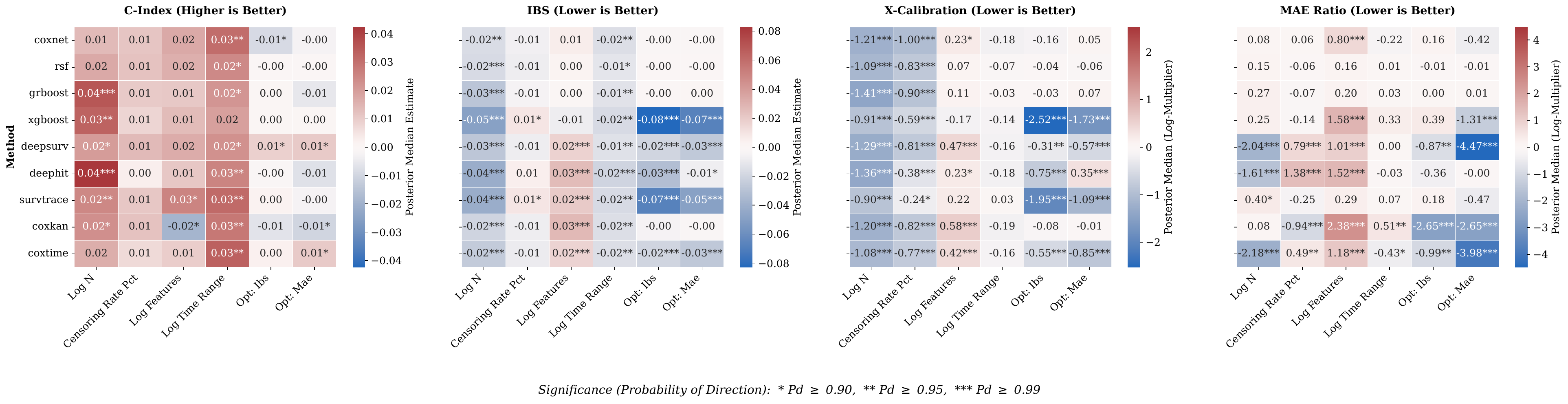}
    
    \caption{Probability of Direction ($pd$) heatmap. Impact of dataset variables and model selection strategy on model performance for all metrics. Cells depict effect strengths. Higher is better for C-index. Lower is better for the rest.}
    \label{fig:pd_heatmaps}
\end{figure}

\begin{figure}[htbp]
    \centering
    \includegraphics[width=\linewidth]{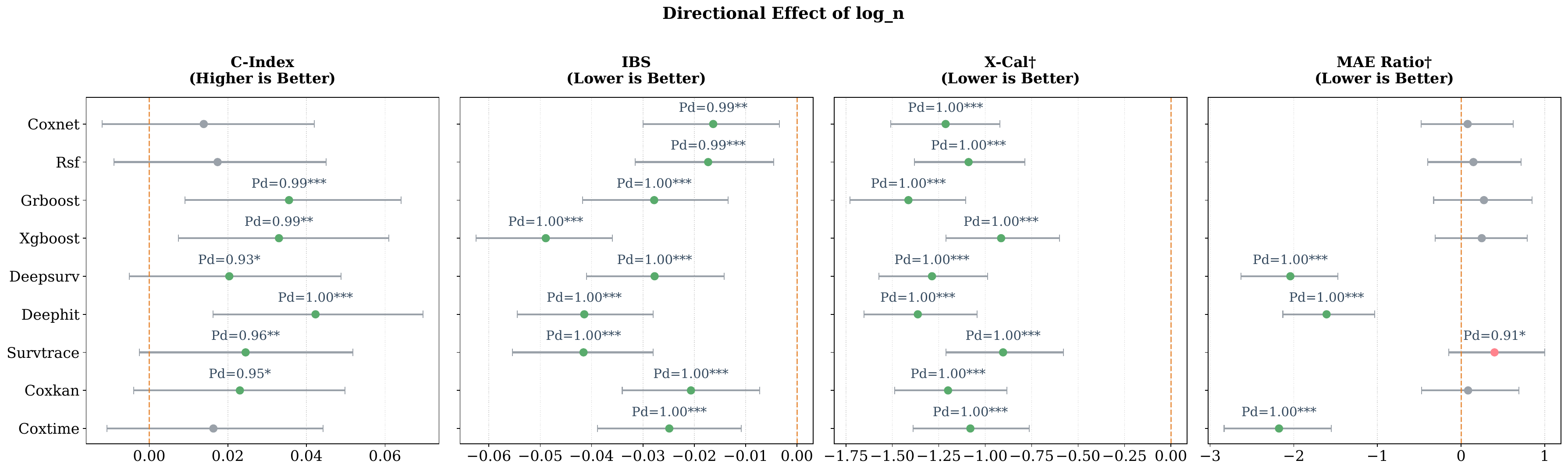}
    
    \caption{Effects of changing the Log-Sample size (Log N) by one standard deviation for each method on target metrics. Significant ($pd$ > 0.9) results are colored. Red: detrimental effect, green: beneficial effect. A dagger ($\dagger$) denotes metrics that are reported on a log-multiplier scale.}
    \label{fig:log_n_pd}
\end{figure}

\begin{figure}[htbp]
    \centering
    \includegraphics[width=0.4\linewidth]{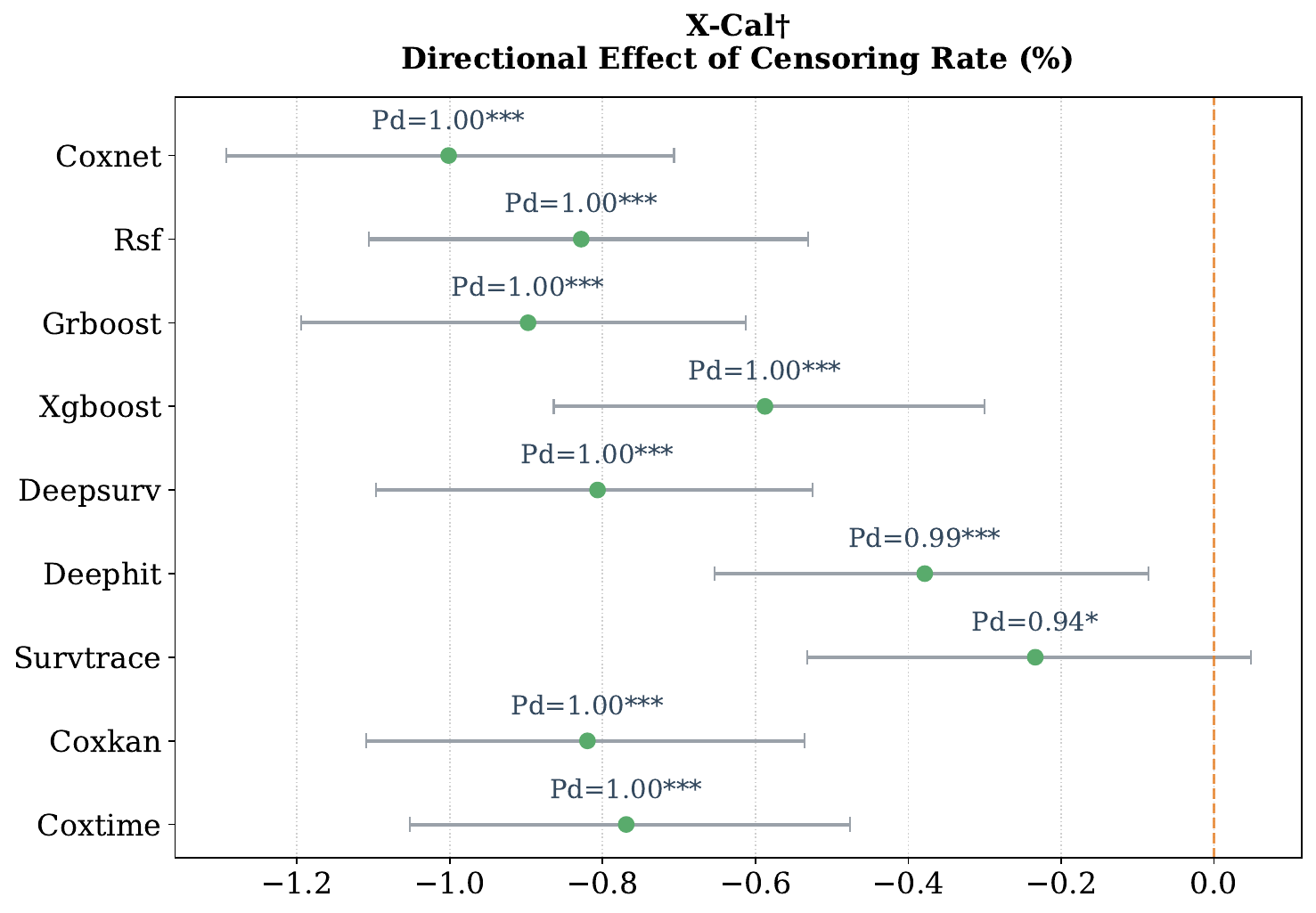}
    \caption{Effects of changing the Censoring Rate (\%) by one standard deviation for each method on target X-CAL. Significant ($pd$ > 0.9) results are colored. Red: detrimental effect, green: beneficial effect. A dagger ($\dagger$) denotes that X-CAL change is reported on a log-multiplier scale.}
    \label{fig:censoring_rate_pd}
\end{figure}

\newpage

\clearpage
\section{Supplementary Results Material}
\label{app:suppl_material}
\subsection{General Supplementary Material}
\label{app:general_suppl_material}

\subsubsection{Heatmaps}

\begin{figure}[H]
    \centering
    \includegraphics[width=1\linewidth]{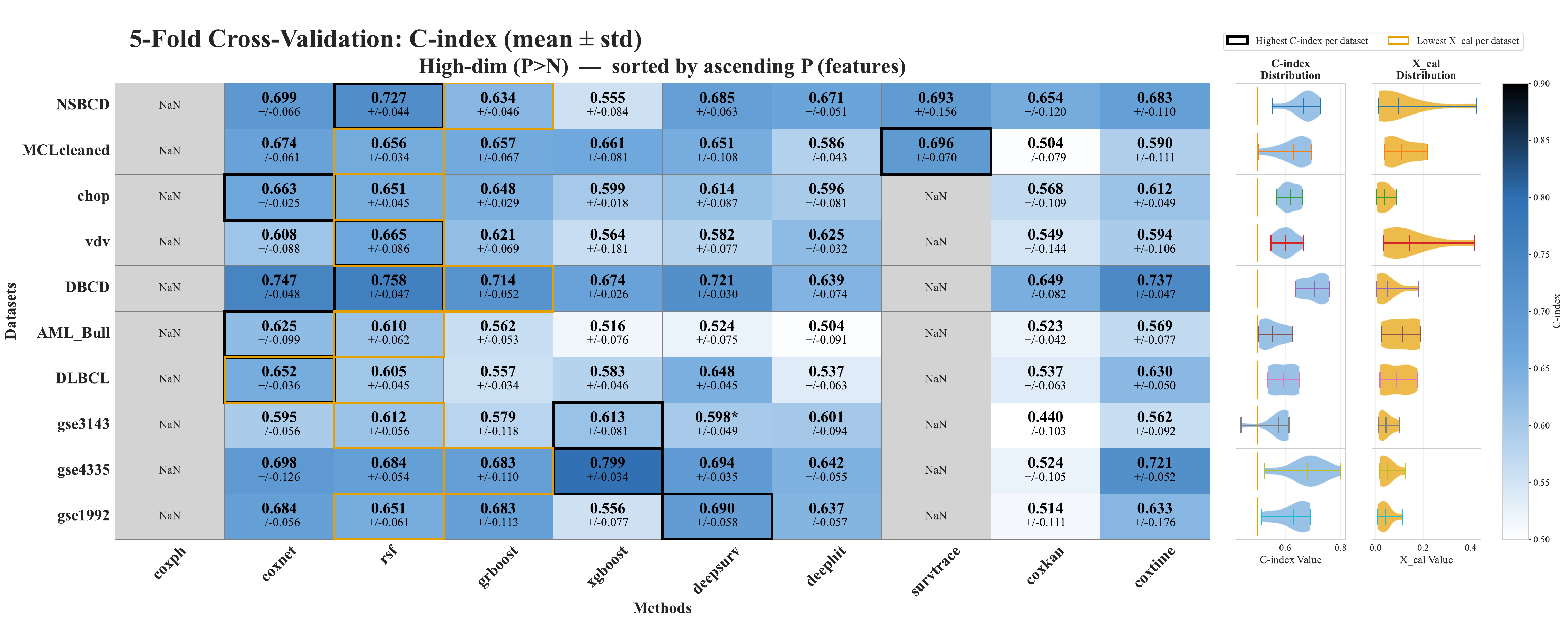}
    \caption{
        \textbf{5-Fold cross-validated Concordance Index (C-index).} Each cell reports the mean~$\pm$~standard deviation across five folds for a given method--dataset combination. The figure shows datasets where $P > N$ (high-dimensional) sorted by ascending feature count~$P$. Cell colour intensity follows a sequential blue scale shared across both panels, ranging from $0.5$ (random discrimination) to $0.9$ (near-perfect discrimination). Gray cells labeled \textit{NaN} indicate that the experiment failed or ran but produced no valid result on all five folds. Cells marked with '*' are based on a subset of the five folds, because at least one fold failed. The C-index violin plots (right) show the cross-method distribution per dataset. A bold black border highlights the highest C-index per dataset and a red border highlights the best X-CAL score per dataset.
    }
    \label{fig:kfold_cindex_heatmap_highdim}
\end{figure}

\begin{figure}[H]
    \centering
    \includegraphics[width=1\linewidth]{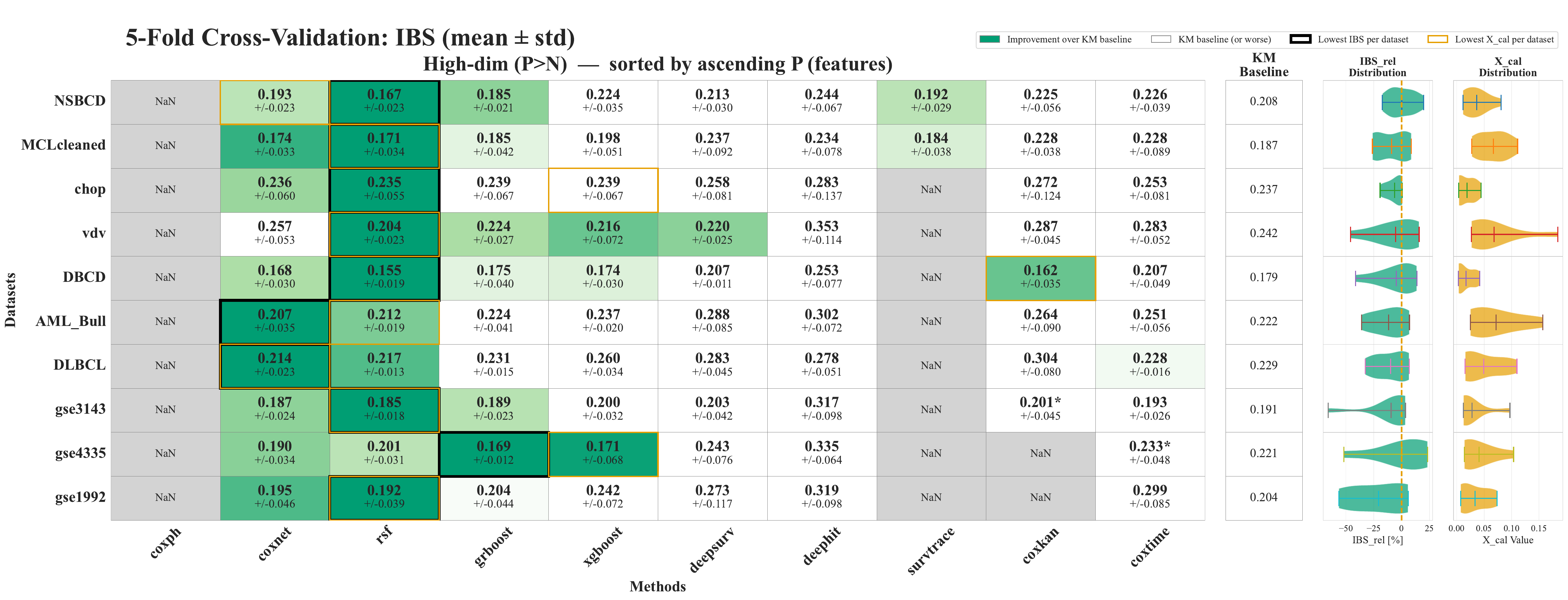}
    \caption{
    \textbf{5-Fold cross-validated Integrated Brier Score (IBS).}
        Each cell reports the mean~$\pm$~standard deviation across five folds for a given method-dataset combination. The figure shows datasets where $P > N$ (high-dimensional) sorted by ascending feature count~$P$. Green shading indicates improvement over the Kaplan-Meier (KM) baseline IBS; intensity scales with the relative gain, with the darkest green corresponding to the best-performing method per dataset. White cells indicate performance at or below the KM baseline. Gray cells labeled \textit{NaN} indicate that the experiment failed or ran but produced no valid result on all five folds. Cells marked with '*' are based on a subset of the five folds, because at least one fold failed. The $\text{IBS}_\text{rel}$ violin plots (right) show the per-dataset distribution of relative improvement over KM across methods; the vertical dashed line marks zero improvement. The \textit{KM Baseline} column reports the dataset-level KM IBS for reference.
    }
    \label{fig:kfold_ibs_heatmap_highdim}
\end{figure}

\begin{figure}
    \centering
    \includegraphics[width=1\linewidth]{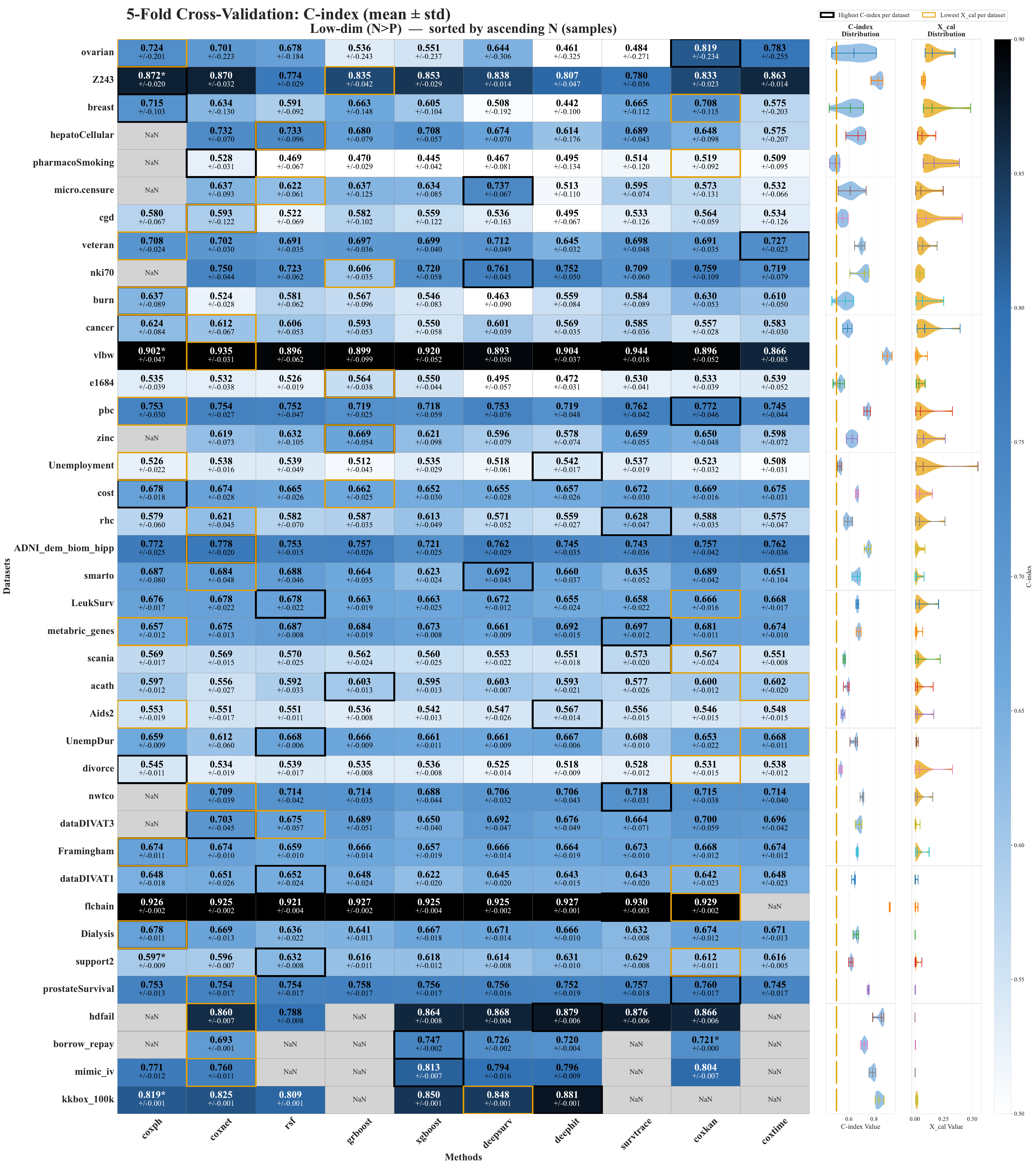}
    \caption{
        \textbf{5-Fold cross-validated Concordance Index (C-index).} Each cell reports the mean~$\pm$~standard deviation across five folds for a given method-dataset combination. The figure shows datasets where $N > P$ (low-dimensional) sorted by ascending sample size~$N$. Cell colour intensity follows a sequential blue scale shared across both panels, ranging from $0.5$ (random discrimination) to $0.9$ (near-perfect discrimination). Gray cells labeled \textit{NaN} indicate that the experiment failed or ran but produced no valid result on all five folds. Cells marked with '*' are based on a subset of the five folds, because at least one fold failed. The C-index violin plots (right) show the cross-method distribution per dataset. A bold black border highlights the highest C-index per dataset and a red border highlights the best X-CAL score per dataset.
    }
    \label{fig:kfold_cindex_heatmap_lowdim}
\end{figure}

\begin{figure}
    \centering
    \includegraphics[width=1\linewidth]{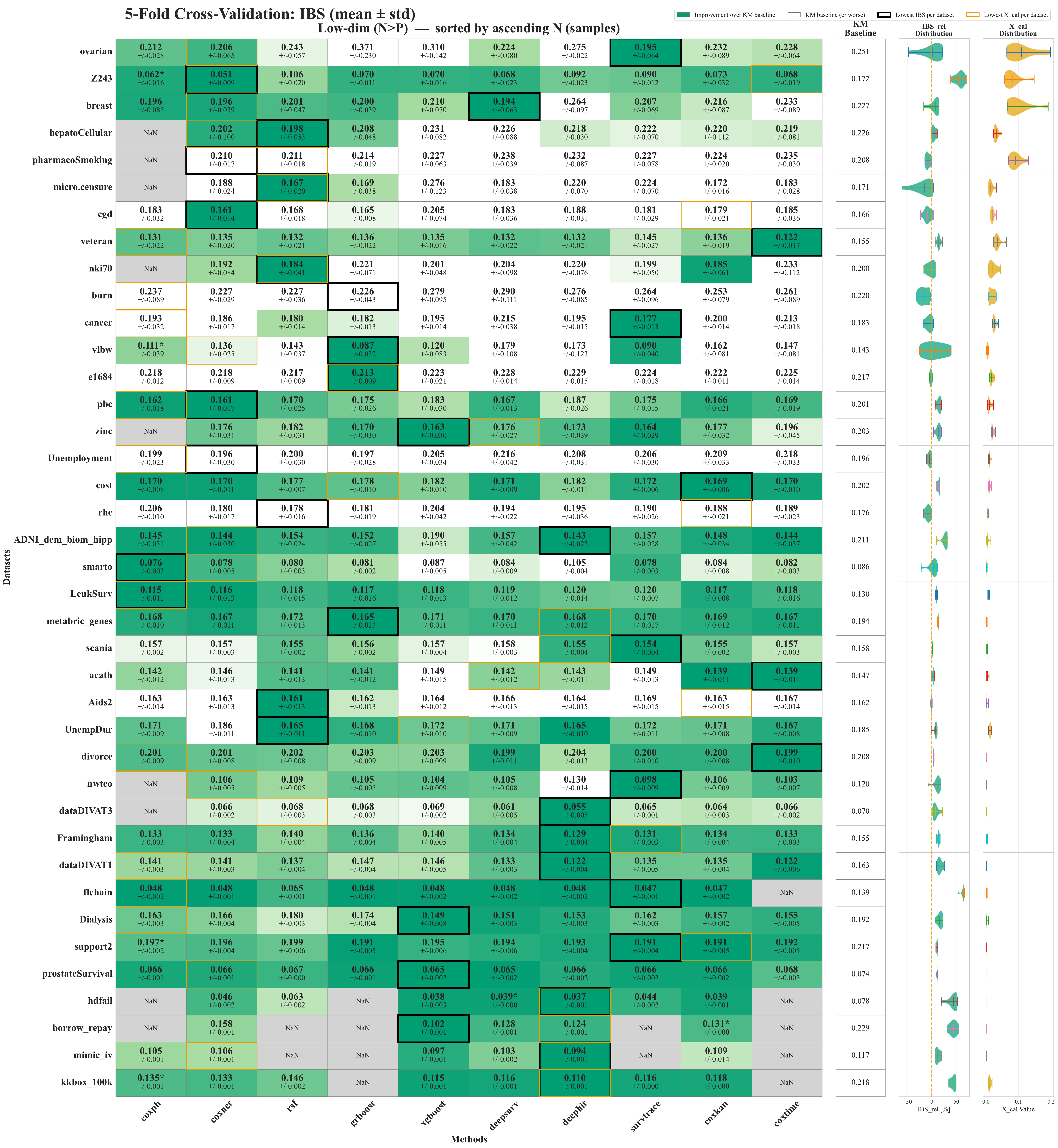}
    \caption{
    \textbf{5-Fold cross-validated Integrated Brier Score (IBS).}
        Each cell reports the mean~$\pm$~standard deviation across five folds for a given method-dataset combination. The figure shows datasets where $N > P$ (low-dimensional) sorted by ascending sample size~$N$. Green shading indicates improvement over the Kaplan-Meier (KM) baseline IBS; intensity scales with the relative gain, with the darkest green corresponding to the best-performing method per dataset. White cells indicate performance at or below the KM baseline. Gray cells labeled \textit{NaN} indicate that the experiment failed or ran but produced no valid result on all five folds. Cells marked with '*' are based on a subset of the five folds, because at least one fold failed. The $\text{IBS}_\text{rel}$ violin plots (right) show the per-dataset distribution of relative improvement over KM across methods; the vertical dashed line marks zero improvement. The \textit{KM Baseline} column reports the dataset-level KM IBS for reference.
    }
    \label{fig:kfold_ibs_heatmap_lowdim}
\end{figure}

\begin{figure}
    \centering
    \includegraphics[width=1\linewidth]{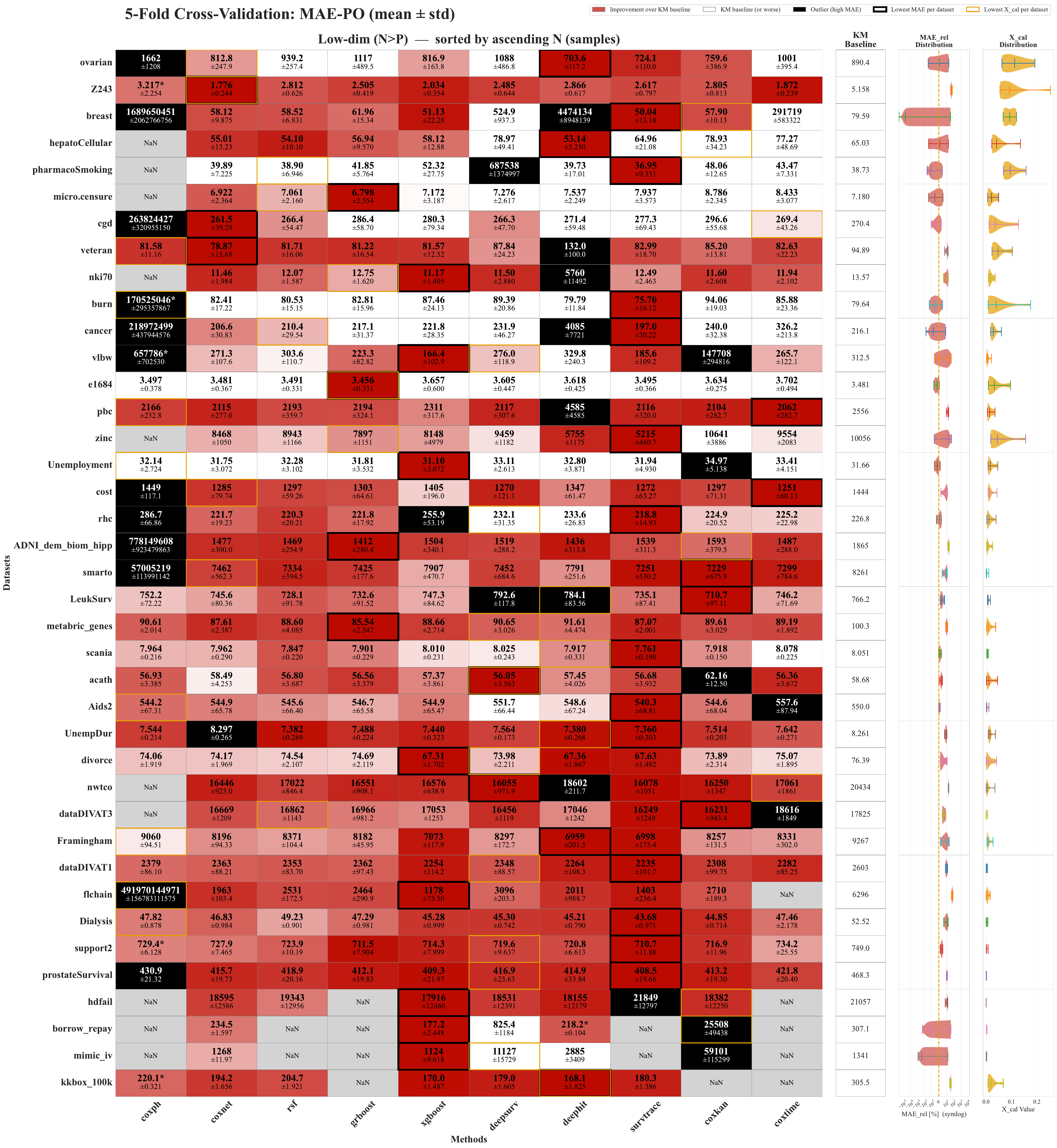}
    \caption{
        \textbf{5-Fold cross-validated Mean Absolute Error (MAE-PO).}
        Each cell reports the mean~$\pm$~standard deviation across five folds for a given method-dataset combination. The figure shows datasets where $N > P$ (low-dimensional) sorted by ascending sample size~$N$. Red shading indicates improvement over the Kaplan-Meier (KM) baseline MAE-PO across a dataset. Black cells mark outliers, defined per dataset row as values exceeding $Q_3 + 1.5 \cdot \text{IQR}$ of the raw MAE distribution across methods. White cells indicate performance at or below the KM baseline. Gray cells labeled \textit{NaN} indicate that the experiment failed or ran but produced no valid result on all five folds. Cells marked with '*' are based on a subset of the five folds, because at least one fold failed. The $MAE\_rel$ violin plots show the relative-improvement distribution per dataset on a log-scale; outliers (black cells) are excluded to prevent extreme values from distorting the scale. The \textit{KM Baseline} column reports the dataset-level KM MAE-PO for reference. A bold black border highlights the lowest MAE per dataset and a red border
        highlights the best X-CAL score per dataset.
    }
    \label{fig:kfold_mae_heatmap_lowdim}
\end{figure}

\begin{figure}
    \centering
    \includegraphics[width=1\linewidth]{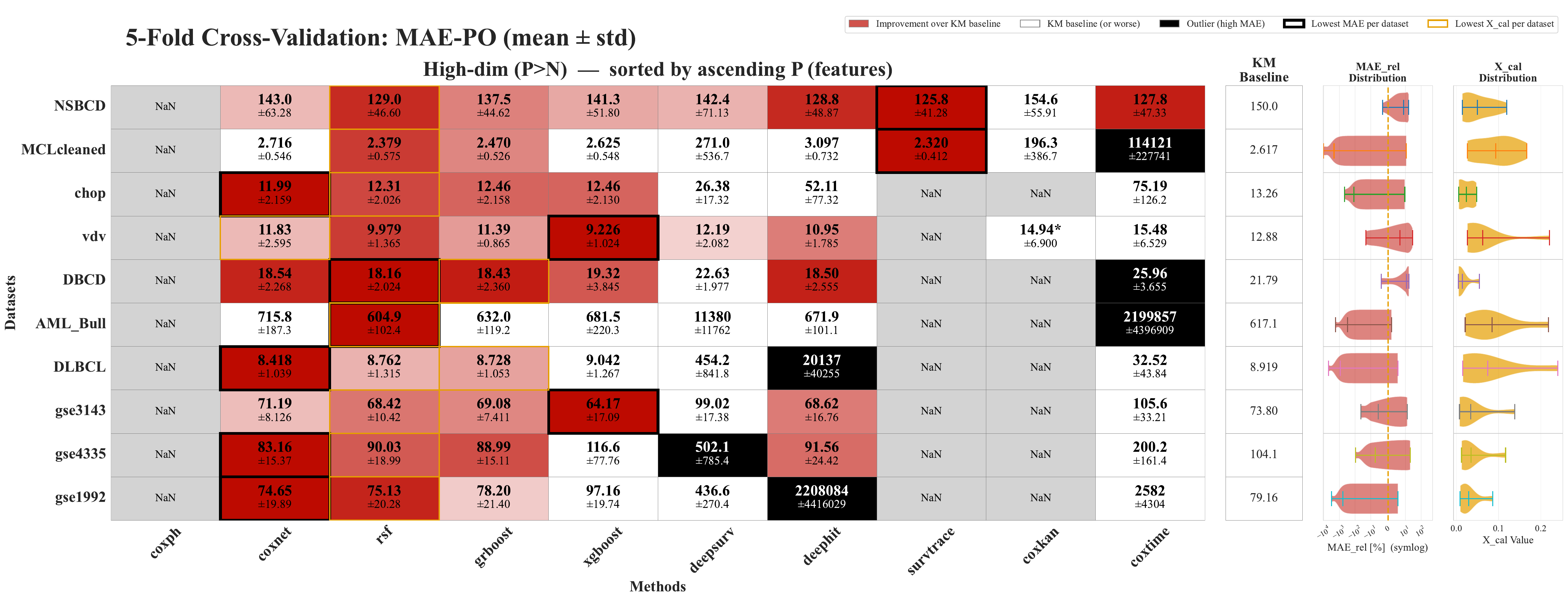}
    \caption{
        \textbf{5-Fold cross-validated Mean Absolute Error (MAE-PO).}
        Each cell reports the mean~$\pm$~standard deviation across five folds for a given method-dataset combination. The figure shows datasets where $P > N$ (high-dimensional) sorted by ascending feature count~$P$. Red shading indicates improvement over the Kaplan-Meier (KM) baseline MAE-PO across a dataset. Black cells mark outliers, defined per dataset row as values exceeding $Q_3 + 1.5 \cdot \text{IQR}$ of the raw MAE distribution across methods. White cells indicate performance at or below the KM baseline. Gray cells labeled \textit{NaN} indicate that the experiment failed or ran but produced no valid result on all five folds. Cells marked with '*' are based on a subset of the five folds, because at least one fold failed. The $MAE\_rel$ violin plots show the relative-improvement distribution per dataset on a log-scale; outliers (black cells) are excluded to prevent extreme values from distorting the scale. The \textit{KM Baseline} column reports the dataset-level KM MAE-PO for reference. A bold black border highlights the lowest MAE per dataset and a red border
        highlights the best X-CAL score per dataset.
    }
    \label{fig:kfold_mae_heatmap_highdim}
\end{figure}

\clearpage
\subsubsection{Ranks}
\begin{figure}[H]
    \centering
    \includegraphics[width=1\linewidth]{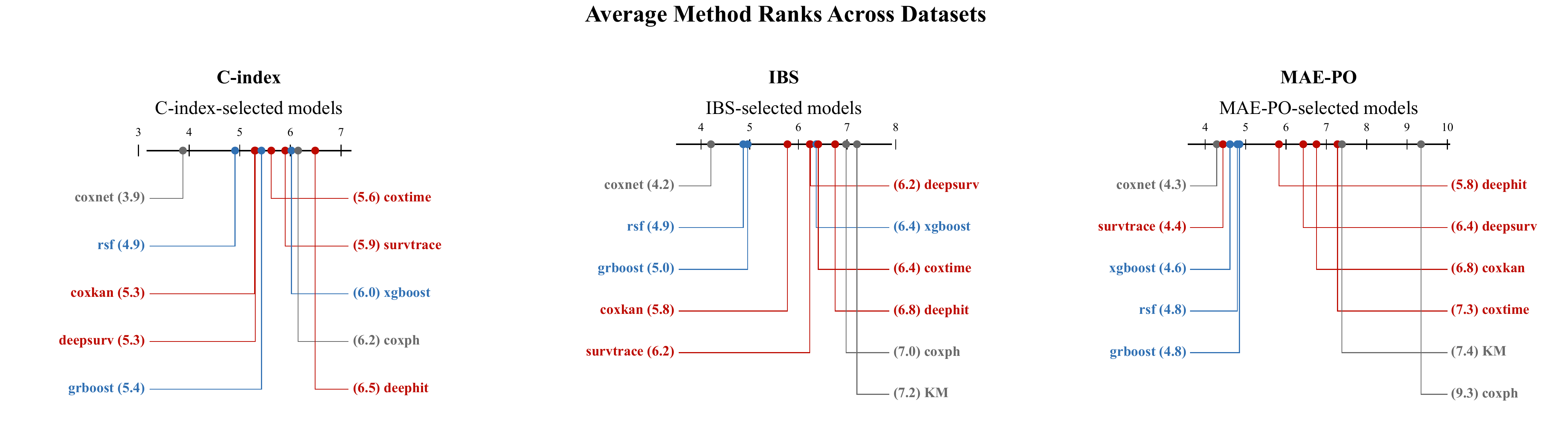}
    \caption{Average Ranks Across Datasets for all three model selection metrics}
    \label{fig:average_rank_cindex_ibs_mae}
\end{figure}

\subsubsection{Performance -- Applicability trade-off}
\begin{figure}[H]
    \centering
    \includegraphics[width=\textwidth]{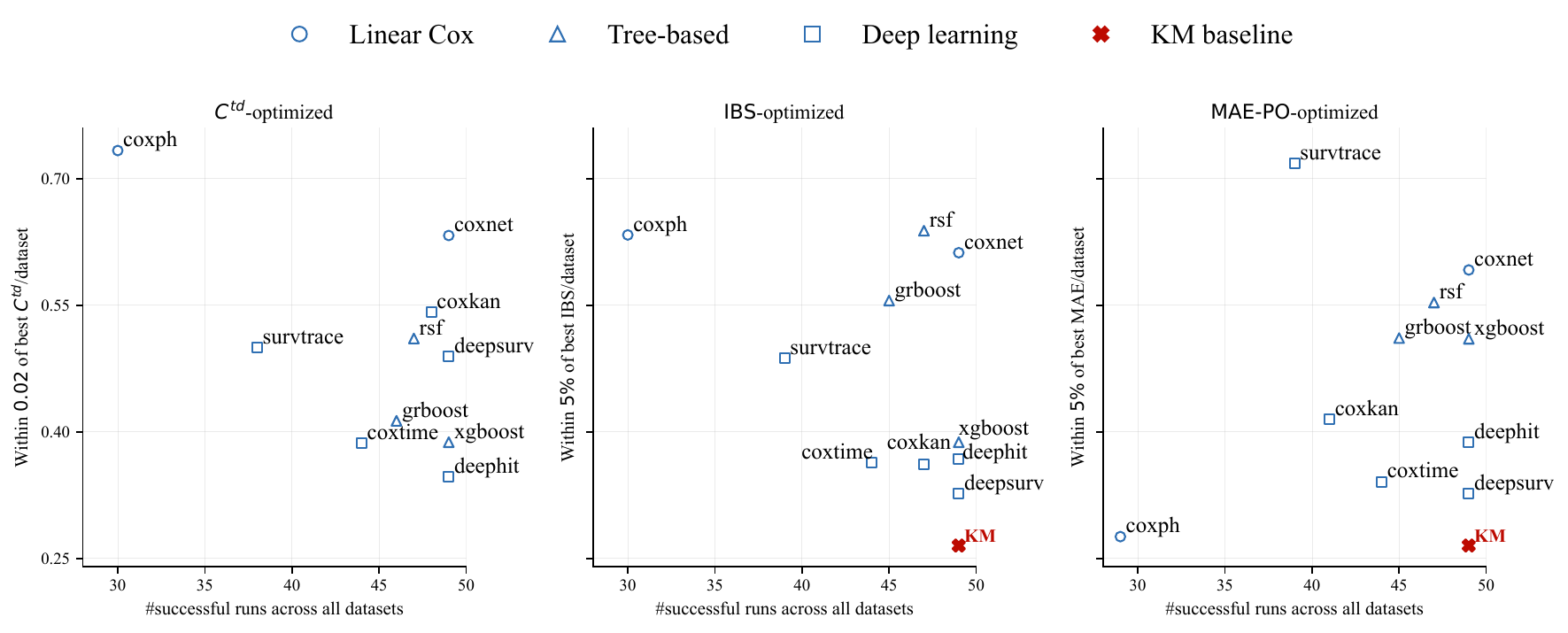}
    \caption{Performance–applicability trade-off across model families for all three model selection criteria. X-axis shows the number of successful runs across all datasets and the y-axis denotes the fraction of how often a method lands within a fixed threshold of the best performing method on that dataset. The threshold for C-index (left) is set to absolute 0.02. As IBS and MAE-PO values are dataset dependent, the threshold is set to 5\% -- meaning that if a method achieved a score within 5\% of the best performing dataset, it will be included in the fraction.}
    \label{fig:tradeoff}
\end{figure}

\clearpage
\subsection{Runtime Analysis}
\label{app:runtime_analysis}
Runtime in our benchmark is hard to disentangle, it is dependent on: (i) the model selection criterion, which affects the number of training epochs through early stopping, (ii) the size of each method’s hyperparameter grid, and (iii) the cost and implementation details of the individual evaluation metrics. Consequently, absolute runtime should not be interpreted as intrinsic algorithmic efficiency.

Our primary objective was to analyze how model selection criteria and dataset characteristics affect predictive performance, rather than benchmark computational efficiency. Within the successfully completed experiments, all methods were evaluated under identical hardware and runtime budgets, allowing controlled comparisons of model selection criteria.

To demonstrate the difficulty in disentangling runtime, we report runtime per hyperparameter (HP) configuration for the intersection of successfully completed method-dataset pairs. The results show substantial runtime differences both across methods and optimization criteria, illustrating that computational cost depends not only on the learning architecture but also on the selected evaluation objective.

\begin{table}[h]
\centering
\caption{Mean and median runtime (s) per HP configuration (C-index optimization).}
\label{tab:runtime-cindex}
\begin{tabular}{lrr}
\hline
Method & Mean (s) & Median (s) \\
\hline
CoxNet    & 0.540  & 0.041 \\
GrBoost   & 2.338  & 0.186 \\
RSF       & 0.414  & 0.245 \\
XGBoost   & 0.356  & 0.269 \\
DeepSurv  & 1.019  & 0.459 \\
DeepHit   & 1.613  & 1.232 \\
CoxKAN    & 2.131  & 1.370 \\
SurvTRACE & 11.119 & 4.154 \\
CoxTime   & 11.868 & 5.556 \\
\hline
\end{tabular}
\end{table}

We additionally compare the mean total runtime (s) for identical method-dataset pairs across different model selection criteria to demonstrate that runtime is influenced by the chosen optimization metric. Together, these show that deep models are slower in our setup both per evaluation and in total, while for flexible tree/boosting models, total cost additionally depends on which optimization metric is used.
\begin{table}[H]
\centering
\caption{Mean runtime (s) per model-selection criterion.}
\label{tab:runtime-selection-criterion}
\begin{tabular}{lrrr}
\hline
Method & C-index (s) & IBS (s) & MAE-PO (s) \\
\hline
CoxPH     & 12.5    & 13.8    & 14.1    \\
CoxNet    & 334.1   & 273.8   & 484.5   \\
RSF       & 361.8   & 418.3   & 1,158.3 \\
XGBoost   & 401.2   & 3,335.6 & 1,197.4 \\
SurvTRACE & 802.7   & 724.0   & 547.8   \\
GrBoost   & 628.0   & 592.2   & 998.8   \\
DeepSurv  & 953.6   & 1,488.1 & 1,055.3 \\
CoxTime   & 1,942.4 & 1,786.0 & 2,058.7 \\
CoxKAN    & 1,315.5 & 2,082.6 & 2,839.3 \\
DeepHit   & 4,073.8 & 5,197.8 & 5,683.4 \\
\hline
\end{tabular}
\end{table}

\clearpage
\subsection{Failure Analysis}
\label{app:failure_analysis}
\textbf{Methods show inherent limitations.} 
Fig. \ref{fig:failure_rate_matrix} shows the failure rate for all experiments broken down by dataset regime and method, illustrated here for $C^{td}$ model selection; results under IBS and MAE-PO selection are qualitatively similar. We mark a dataset–method pair as failed if any fold did not produce results. The failures exhibit systematic structure. The failures are neither uniformly distributed across the matrix nor confined to individual methods. Instead, they concentrate in the large-$N$ and high-$P$ regimes, revealing where current survival analysis architectures reach their practical limits. The following subsections show the broken down analyses of the failure reasons runtime, memory and mathematical errors. 

\subsubsection{Runtime}
We distinguish between symmetric runtime failures and asymmetric runtime failures. The symmetric runtime failure matrix highlights systematic scalability limitations. This is consistent with preliminary experiments before imposing the runtime budget: on hdfail (N > 54k), GrBoost required ~65h on a 32-core CPU with 128 GB RAM, compared to ~7 minutes for CoxNet (+6.4\% C-index) and ~2.25 hours for DeepSurv (+7.4\% C-index). This suggests that these failures reflect practical scalability limitations rather than the chosen budget. 

\begin{table}[ht]
\centering
\begin{tabular}{lll}
\hline
Method & Regime & Share \\
\hline
GrBoost & $N \geq 10\mathrm{k}$, low-dim & 67\% (4/6) \\
RSF     & $N \geq 10\mathrm{k}$, low-dim & 33\% (2/6) \\
\hline
\end{tabular}
\end{table}

In contrast, asymmetric runtime failures occur only for some model selection criteria. These are primarily explained by different optimization trajectories and metric evaluation costs. Consequently, increasing the runtime budget would most likely resolve many of these failures. A detailed analysis of the runtime in general is provided in \ref{app:runtime_analysis}. 
\\

\begin{table}[ht]
\centering
\begin{tabular}{lll}
\hline
Method & Regime & Share \\
\hline
CoxKAN    & $100 \leq N < 1\mathrm{k}$, high-dim & 59\% (10/17) \\
CoxKAN    & $N \geq 10\mathrm{k}$, low-dim      & 23\% (4/17) \\
CoxKAN    & $N < 100$, high-dim                 & 6\% (1/17) \\
DeepHit   & $N \geq 10\mathrm{k}$, low-dim      & 6\% (1/17) \\
SurvTRACE & $N \geq 10\mathrm{k}$, low-dim      & 6\% (1/17) \\
\hline
\end{tabular}
\end{table}

\subsubsection{Memory}
Memory failures, in contrast, cannot be resolved by extending the runtime budget. They occur systematically for the following method–regime combinations: \\
\begin{table}[H]
\centering
\begin{tabular}{lll}
\hline
Method & Regime & Share \\
\hline
SurvTRACE & high-dim ($P > 1\mathrm{k}$)              & 53\% (8/15) \\
SurvTRACE & $N \geq 10\mathrm{k}$, low-dim             & 13\% (2/15) \\
CoxTime   & $N \geq 10\mathrm{k}$, low-dim             & 27\% (4/15) \\
CoxTime   & $1\mathrm{k} \leq N < 10\mathrm{k}$, low-dim & 7\% (1/15) \\
\hline
\end{tabular}
\end{table}

Memory failures are concentrated in CoxTime and SurvTRACE and occur consistently across all optimization targets within specific dataset regimes. They therefore likely reflect architecture-specific scalability limitations rather than the optimization objective. We observed increasing memory usage up to the available capacity, but did not collect detailed memory traces to identify the underlying causes.

\newpage
\subsubsection{Mathematical Errors}
Only unregularized CoxPH (baseline) exhibits mathematical failures (e.g., ill-conditioned matrices), indicating inherent numerical stability limitations rather than resource constraints, particularly in challenging high-dim and large-N regimes.

\begin{figure}
    \centering
    \includegraphics[width=1\linewidth]{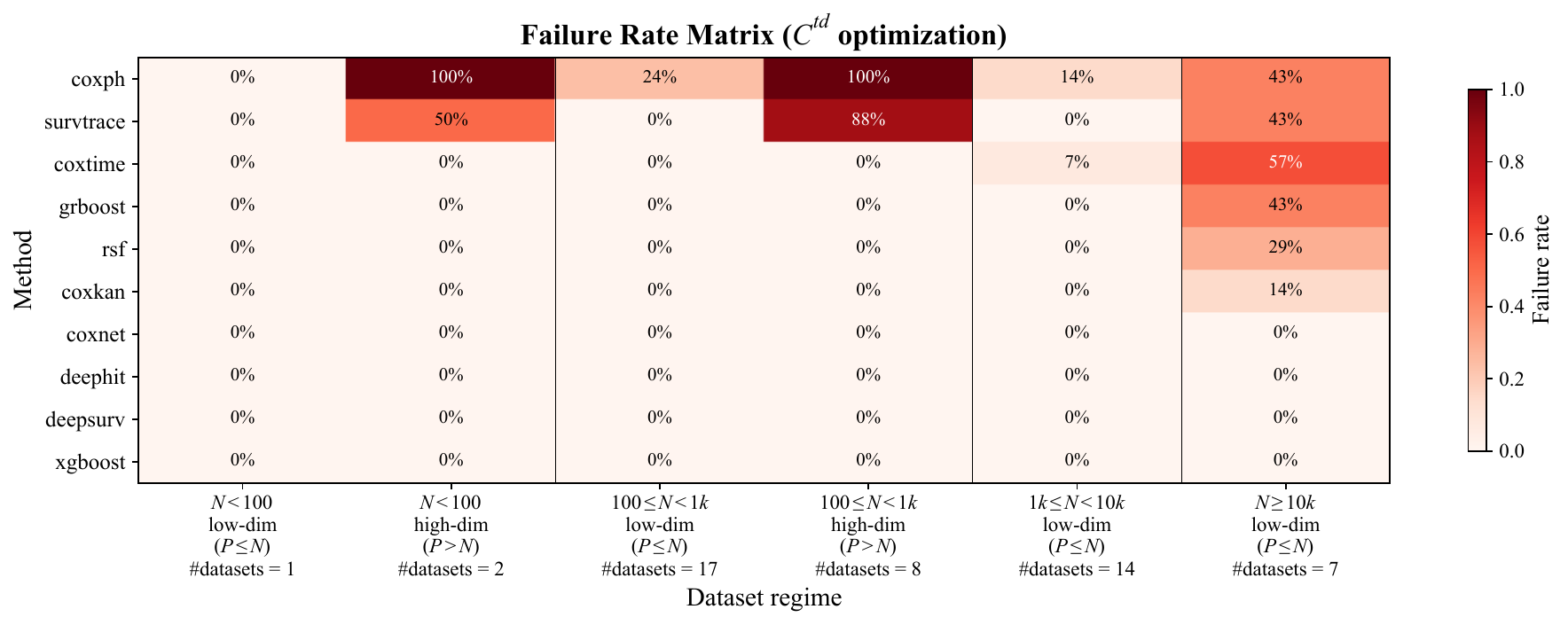}
    \caption{Failure rate matrix: fraction of datasets where a method fails under $C^{td}$-optimized hyperparameters. Structured by observation count ($N$) and feature count ($P$).}
    \label{fig:failure_rate_matrix}
\end{figure}

\subsection{Robustness Analysis on the Common Intersection of Successful Runs}
\label{app:robustness_common_intersection}

\begin{figure}[htbp]
    \centering
    \includegraphics[width=\linewidth]{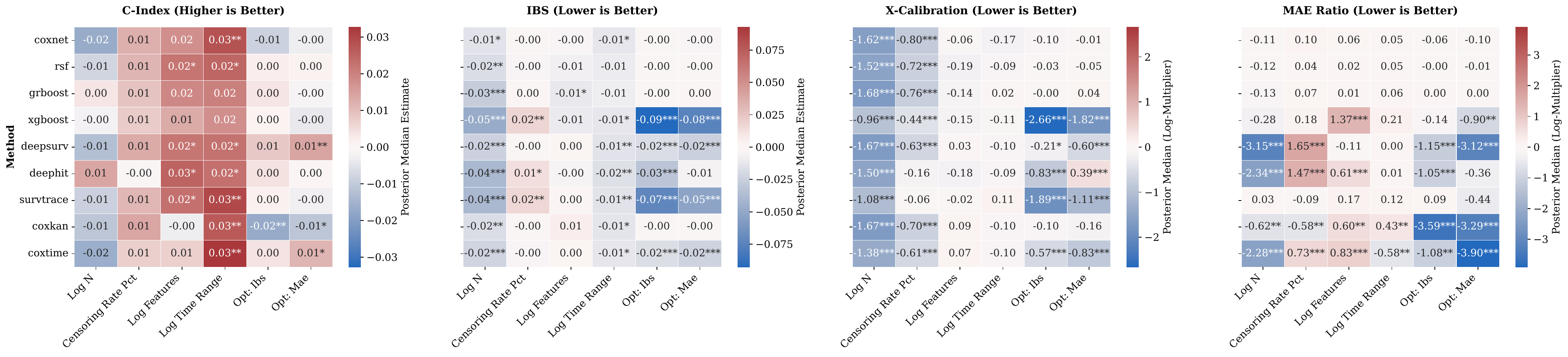}
    
    \caption{Probability of Direction ($pd$) heatmap on the common intersection of successful runs. Impact of dataset variables and model selection strategy on model performance for all metrics. Cells depict effect strengths. Higher is better for C-index. Lower is better for the rest.}
    \label{fig:pd_heatmaps_intersection}
\end{figure}

In Fig. \ref{fig:failure_rate_matrix}, we see that some methods have systematical errors and could not run on certain datasets. This obviously might create a survivorship bias in our analysis. Here, we show that our Bayesian GLMM actually produced reliable results, by re-creating the Probability of Direction and median effect size analysis on the common intersection of successful runs. The median effect strengths with their probabilities of direction can be seen in Fig. \ref{fig:pd_heatmaps_intersection}. All our main findings from the main body of the manuscript can be inferred from this balanced subset equally. Compare against Fig. \ref{fig:pd_heatmaps}.

\vspace{2mm}

\clearpage
\subsection{\texorpdfstring{\protect\circnumA{A.2}}{A.2} Supplementary Material}
\paragraph{Rank Shifts When Switching Target}
\leavevmode \par
\newcommand{\cellcolorval}[1]{%
\ifdim #1 pt > 0.001pt
    \cellcolor{LightGreen}{#1}%
\else
    \ifdim #1 pt < -0.001pt
        \cellcolor{LightRed}{#1}%
    \else
        #1%
    \fi
\fi
}

\begin{table}[H]
\centering
\caption{
Median paired effect of switching model selection targets.
Each panel shows $\Delta (\cdot) = (\cdot)^{B} - (\cdot)^{A}$ ($A-B$ for X\text{-}CAL, as lower is better), where the superscript denotes the model selection criterion. All deltas are oriented such that $\uparrow$ indicates improvement. $C^{td}$ is reported on its standard scale in $[0,1]$ (higher is better). $IBS\_{\mathrm{rel}}$ and $MAE\_{\mathrm{rel}}$ denote relative improvement over the Kaplan–Meier baseline, unitless but interpretable as percentage gains. X\text{-}CAL (lower is better) is inverted here such that positive values indicate improved calibration.}

\label{tab:selection_switch_effects_descriptive}
\vspace{0.5em}
\resizebox{\linewidth}{!}{%
\begin{tabular}{lrrrr|rrrr|rrrr}
\toprule
& \multicolumn{4}{c}{\boldmath$ (\cdot)^{C^{td}} \rightarrow (\cdot)^{IBS} $}
& \multicolumn{4}{c}{\boldmath$(\cdot)^{C^{td}} \rightarrow (\cdot)^{MAE}$}
& \multicolumn{4}{c}{\boldmath$(\cdot)^{IBS} \rightarrow (\cdot)^{MAE}$} \\
\cmidrule(lr){2-5}
\cmidrule(lr){6-9}
\cmidrule(lr){10-13}
\textbf{Method}
& $\Delta C^{td}\uparrow$ & $\Delta IBS_{\mathrm{rel}}\uparrow$ & $\Delta MAE_{\mathrm{rel}}\uparrow$ & $\Delta XCal\uparrow$

& $\Delta C^{td}\uparrow$ & $\Delta IBS_{\mathrm{rel}}\uparrow$ & $\Delta MAE_{\mathrm{rel}}\uparrow$ & $\Delta XCal\uparrow$

& $\Delta C^{td}\uparrow$ & $\Delta IBS_{\mathrm{rel}}\uparrow$ & $\Delta MAE_{\mathrm{rel}}\uparrow$ & $\Delta XCal\uparrow$ \\
\midrule

coxnet
& \cellcolorval{0.000} & \cellcolorval{0.000} & \cellcolorval{0.000} & \cellcolorval{0.000}
& \cellcolorval{0.000} & \cellcolorval{0.000} & \cellcolorval{0.008} & \cellcolorval{0.000}
& \cellcolorval{0.000} & \cellcolorval{-0.001} & \cellcolorval{0.005} & \cellcolorval{0.000} \\

rsf
& \cellcolorval{0.000} & \cellcolorval{0.003} & \cellcolorval{0.002} & \cellcolorval{0.000}
& \cellcolorval{-0.001} & \cellcolorval{0.000} & \cellcolorval{0.003} & \cellcolorval{0.000}
& \cellcolorval{0.000} & \cellcolorval{-0.002} & \cellcolorval{0.000} & \cellcolorval{0.000} \\

grboost
& \cellcolorval{0.000} & \cellcolorval{0.001} & \cellcolorval{0.002} & \cellcolorval{0.000}
& \cellcolorval{-0.001} & \cellcolorval{0.000} & \cellcolorval{0.000} & \cellcolorval{0.000}
& \cellcolorval{-0.001} & \cellcolorval{-0.001} & \cellcolorval{0.000} & \cellcolorval{0.000} \\

xgboost
& \cellcolorval{-0.002} & \cellcolorval{0.302} & \cellcolorval{0.008} & \cellcolorval{0.094}
& \cellcolorval{-0.003} & \cellcolorval{0.239} & \cellcolorval{0.100} & \cellcolorval{0.063}
& \cellcolorval{0.000} & \cellcolorval{-0.054} & \cellcolorval{0.098} & \cellcolorval{-0.012} \\

deepsurv
& \cellcolorval{-0.001} & \cellcolorval{0.053} & \cellcolorval{0.028} & \cellcolorval{0.001}
& \cellcolorval{0.002} & \cellcolorval{0.057} & \cellcolorval{51.471} & \cellcolorval{0.002}
& \cellcolorval{0.000} & \cellcolorval{0.004} & \cellcolorval{0.645} & \cellcolorval{0.001} \\

deephit
& \cellcolorval{-0.005} & \cellcolorval{0.132} & \cellcolorval{-0.004} & \cellcolorval{0.005}
& \cellcolorval{-0.009} & \cellcolorval{0.013} & \cellcolorval{0.035} & \cellcolorval{0.000}
& \cellcolorval{-0.005} & \cellcolorval{-0.074} & \cellcolorval{0.024} & \cellcolorval{-0.009} \\

coxtime
& \cellcolorval{-0.002} & \cellcolorval{0.051} & \cellcolorval{0.447} & \cellcolorval{0.002}
& \cellcolorval{0.005} & \cellcolorval{0.087} & \cellcolorval{5.389} & \cellcolorval{0.007}
& \cellcolorval{0.003} & \cellcolorval{0.002} & \cellcolorval{0.300} & \cellcolorval{0.001} \\

survtrace
& \cellcolorval{-0.002} & \cellcolorval{0.179} & \cellcolorval{0.010} & \cellcolorval{0.043}
& \cellcolorval{-0.007} & \cellcolorval{0.129} & \cellcolorval{0.111} & \cellcolorval{0.036}
& \cellcolorval{-0.004} & \cellcolorval{-0.079} & \cellcolorval{0.128} & \cellcolorval{-0.004} \\

coxkan
& \cellcolorval{-0.004} & \cellcolorval{0.006} & \cellcolorval{0.004} & \cellcolorval{0.000}
& \cellcolorval{-0.004} & \cellcolorval{0.000} & \cellcolorval{0.048} & \cellcolorval{0.000}
& \cellcolorval{0.000} & \cellcolorval{0.000} & \cellcolorval{0.028} & \cellcolorval{0.000} \\
\bottomrule
\end{tabular}%
}
\end{table}

\paragraph{Effects of Switching the Model Selection Criterion}
\leavevmode \par
\begin{figure}[H]
    \centering
    \includegraphics[width=1.0\linewidth]{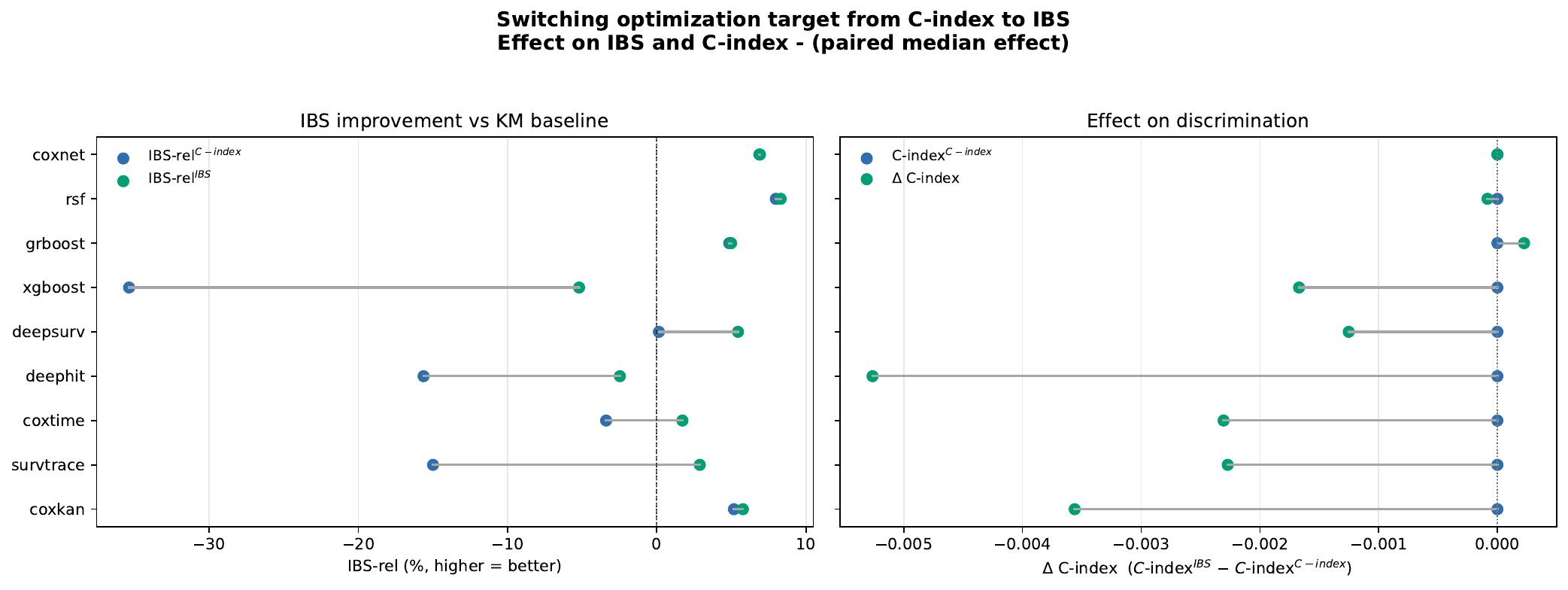}
    \caption{Effect of switching the hyperparameter optimization target from C-index to IBS. Deep learning and boosting methods gain substantially in IBS-rel with negligible C-index change (\emph{free lunch}), while classical methods show little effect. CoxPH is excluded as no tuning is involved.}
    \label{fig:switch_model_selection_criteria}
\end{figure}

\begin{figure}[H]
    \centering
    \includegraphics[width=1.0\linewidth]{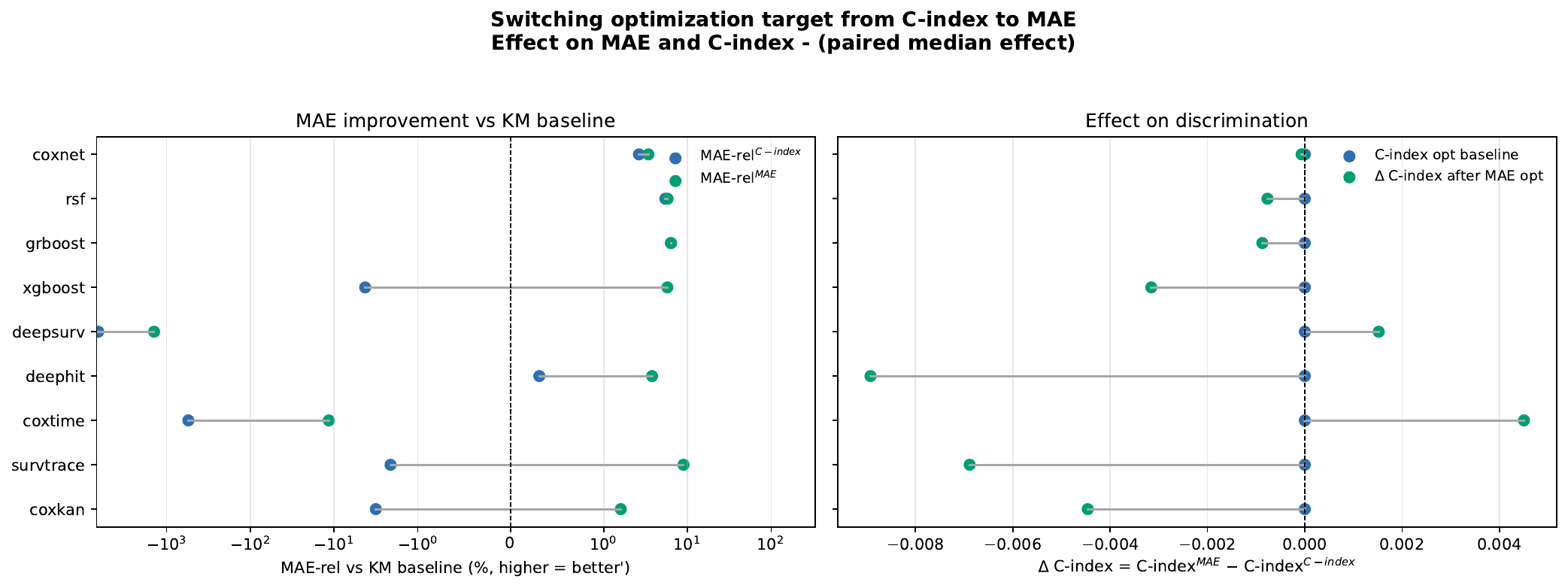}
    \caption{Effect of switching the hyperparameter optimization target from C-index to MAE. MAE improves and becomes more stable across methods, with fewer extreme failures. The C-index remains largely unchanged, indicating that optimizing for time accuracy does not harm discrimination (\emph{free lunch}).}
    \label{fig:cindex_to_mae_cindexEffect}
\end{figure}

\begin{figure}[H]
    \centering
    \includegraphics[width=1.0\linewidth]{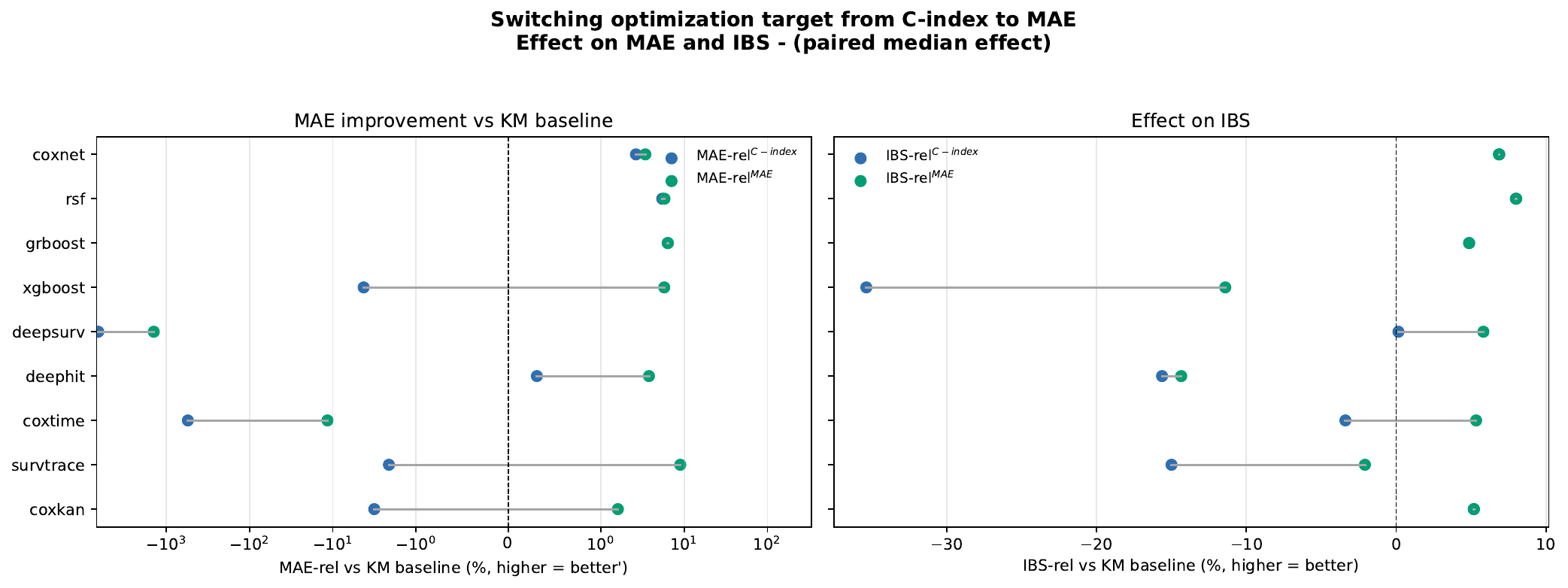}
    \caption{Effect of switching the hyperparameter optimization target from C-index to MAE, evaluated in terms of IBS-rel. Many methods show improved IBS relative to the Kaplan-Meier baseline, indicating better overall predictive accuracy. This demonstrates a \emph{free lunch} effect where optimizing for MAE improves calibration-sensitive metrics without sacrificing discrimination.}
    \label{fig:cindex_to_mae_ibsEffect}
\end{figure}

\begin{figure}[H]
    \centering
    \includegraphics[width=1.0\linewidth]{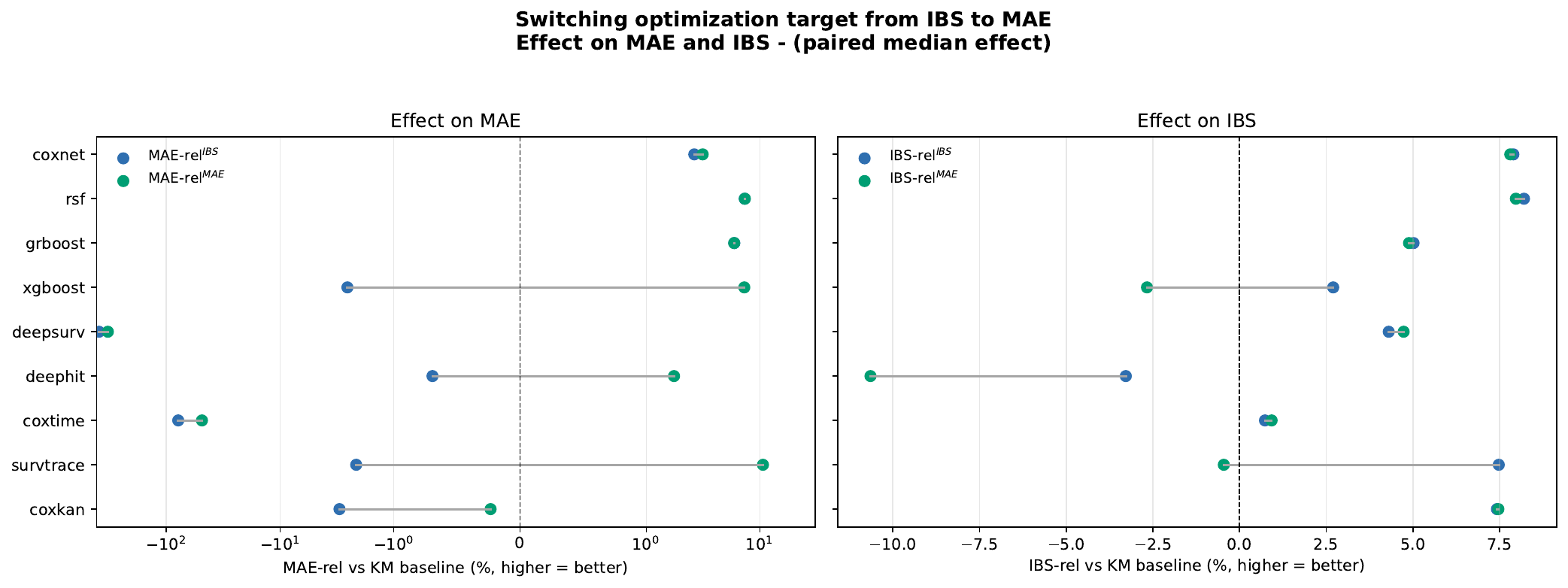}
    \caption{Effect of switching the hyperparameter optimization target from IBS to MAE. MAE improves consistently across most methods, confirming that direct optimization of time accuracy is effective. IBS shows mixed behavior: CoxNet, CoxKAN, DeepSurv and CoxTime benefit (\emph{free lunch}), while boosting and transformer-based methods (e.g. XGBoost, SurvTRACE, DeepHit) on average perform better when directly optimized for IBS. RSF remains unaffected.}
    \label{fig:ibs_to_mae_ibsEffect}
\end{figure}

\begin{figure}[H]
    \centering
    \includegraphics[width=1\linewidth]{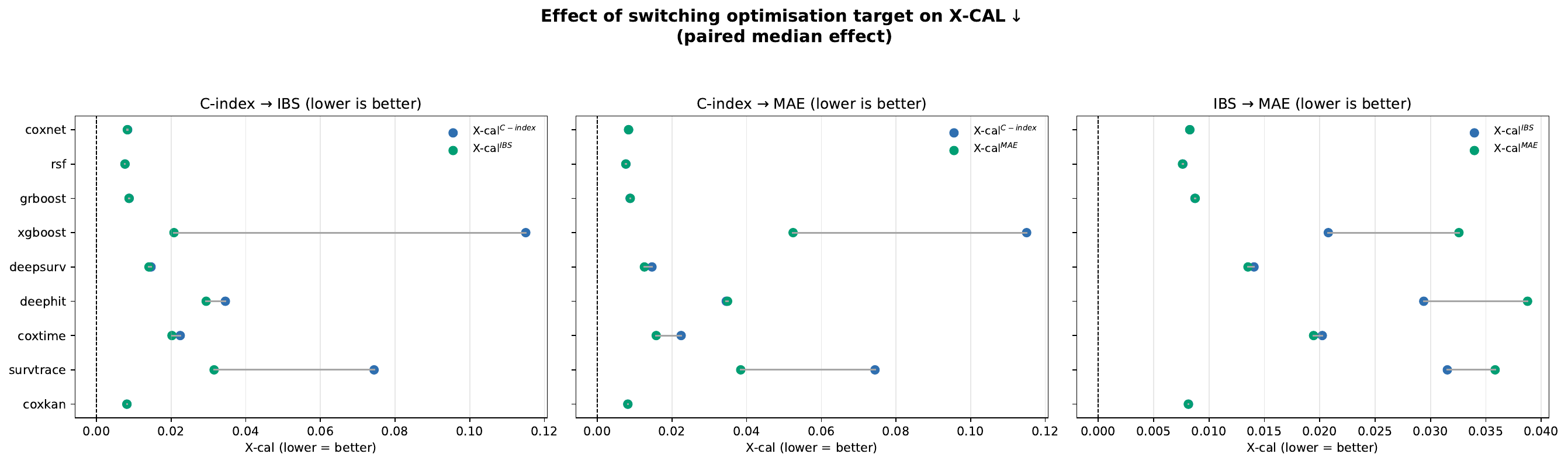}
    \caption{Effect on X-CAL when switching the hyperparameter optimization target. Lower values indicate better calibration. Switching from C-index to IBS yields substantial improvements across all methods, confirming that IBS optimization directly improves calibration. Switching from C-index to MAE also improves calibration for most methods, though typically to a lesser extent. When comparing IBS to MAE optimization, two regimes emerge: CoxKAN, DeepSurv and CoxTime, MAE optimization further improves calibration, whereas for XGBoost, SurvTRACE, and DeepHit, calibration is best when directly optimized via IBS.}
    \label{fig:switch_target_xcal}
\end{figure}

\clearpage
\paragraph{Mean MAE-PO Behaviour}
\leavevmode \par
How mean MAE-PO behaves when switching model selection criterion. 

\begin{table}[H]
\centering
\caption{
\textbf{MAE Improvment and Outlier Analyis.} Relative MAE improvement over the Kaplan-Meier (KM) baseline for different model selection criteria. Positive values indicate improvement over KM. Outliers are computed per dataset across methods using the lower-tail $1.5\cdot IQR$ rule on relative MAE improvement values. The table reports the frequency of outperforming KM ($>$KM), median relative MAE improvement, the proportion of dataset-level failures (\% outliers), and the median degradation among those failures. Lower outlier rates and degradation values indicate more stable behavior.}
\label{tab:mae_rel_summary_all}
\vspace{0.3em}

\scriptsize
\setlength{\tabcolsep}{3.2pt}

\resizebox{\textwidth}{!}{%
\begin{tabular}{lrrrr|rrrr|rrrr}
\toprule

& \multicolumn{4}{c}{\boldmath$C^{td}$-optimized}
& \multicolumn{4}{c}{\boldmath$IBS$-optimized}
& \multicolumn{4}{c}{\boldmath$MAE$-optimized} \\

\cmidrule(lr){2-5}
\cmidrule(lr){6-9}
\cmidrule(lr){10-13}

\textbf{Method}
& \begin{tabular}{@{}c@{}}$>$KM\\(\%)$\uparrow$\end{tabular}
& \begin{tabular}{@{}c@{}}Median\\MAE$_{\mathrm{rel}}$ (\%)$\uparrow$\end{tabular}
& \begin{tabular}{@{}c@{}}Out.\\(\%)$\downarrow$\end{tabular}
& \begin{tabular}{@{}c@{}}Median\\Degr. (\%)$\downarrow$\end{tabular}

& \begin{tabular}{@{}c@{}}$>$KM\\(\%)$\uparrow$\end{tabular}
& \begin{tabular}{@{}c@{}}Median\\MAE$_{\mathrm{rel}}$ (\%)$\uparrow$\end{tabular}
& \begin{tabular}{@{}c@{}}Out.\\(\%)$\downarrow$\end{tabular}
& \begin{tabular}{@{}c@{}}Median\\Degr. (\%)$\downarrow$\end{tabular}

& \begin{tabular}{@{}c@{}}$>$KM\\(\%)$\uparrow$\end{tabular}
& \begin{tabular}{@{}c@{}}Median\\MAE$_{\mathrm{rel}}$ (\%)$\uparrow$\end{tabular}
& \begin{tabular}{@{}c@{}}Out.\\(\%)$\downarrow$\end{tabular}
& \begin{tabular}{@{}c@{}}Median\\Degr. (\%)$\downarrow$\end{tabular} \\

\midrule

coxnet
& 69.4 & 2.63 & \cellcolor{green!20}0.0 & \cellcolor{green!20}0.00
& 71.4 & 2.64 & 2.0 & 892624.20
& \cellcolor{green!20}85.7 & \cellcolor{green!20}8.71 & 2.0 & 0.44 \\

rsf
& 83.0 & 5.44 & \cellcolor{green!20}0.0 & \cellcolor{green!20}0.00
& 85.1 & 7.34 & \cellcolor{green!20}0.0 & \cellcolor{green!20}0.00
& \cellcolor{green!20}89.4 & \cellcolor{green!20}8.14 & \cellcolor{green!20}0.0 & \cellcolor{green!20}0.00 \\

grboost
& \cellcolor{green!20}84.4 & 6.33 & \cellcolor{green!20}0.0 & \cellcolor{green!20}0.00
& \cellcolor{green!20}84.4 & 5.94 & \cellcolor{green!20}0.0 & \cellcolor{green!20}0.00
& \cellcolor{green!20}84.4 & \cellcolor{green!20}6.39 & \cellcolor{green!20}0.0 & \cellcolor{green!20}0.00 \\

xgboost
& 49.0 & -4.25 & \cellcolor{green!20}6.1 & 86.15
& 46.9 & -2.56 & \cellcolor{green!20}6.1 & 434698.87
& \cellcolor{green!20}77.6 & \cellcolor{green!20}9.60 & \cellcolor{green!20}6.1 & \cellcolor{green!20}4.29 \\

deepsurv
& 20.4 & -6560.74 & \cellcolor{green!20}55.1 & 1541743.88
& 16.3 & -391.81 & 57.1 & 2037.14
& \cellcolor{green!20}55.1 & \cellcolor{green!20}3.16 & \cellcolor{green!20}6.1 & \cellcolor{green!20}382.17 \\

deephit
& 51.0 & 0.71 & \cellcolor{green!20}8.2 & 405221.69
& 46.9 & -0.44 & \cellcolor{green!20}12.2 & \cellcolor{green!20}232.46
& \cellcolor{green!20}59.2 & \cellcolor{green!20}4.37 & 20.4 & 934.85 \\

coxtime
& 9.1 & -678.90 & 61.4 & 36704.92
& 18.2 & -79.37 & 36.4 & 3619.42
& \cellcolor{green!20}52.3 & \cellcolor{green!20}0.53 & \cellcolor{green!20}15.9 & \cellcolor{green!20}50.94 \\

survtrace
& 42.1 & -2.12 & \cellcolor{green!20}2.6 & 5.91
& 43.6 & -1.87 & 7.7 & 8.60
& \cellcolor{green!20}87.2 & \cellcolor{green!20}11.47 & \cellcolor{green!20}2.6 & \cellcolor{green!20}3.76 \\

coxkan
& 38.3 & -12.95 & 12.8 & 2272241.02
& 37.0 & -7.20 & 10.9 & 2097.30
& \cellcolor{green!20}61.0 & \cellcolor{green!20}4.29 & \cellcolor{green!20}9.8 & \cellcolor{green!20}6257.43 \\

\midrule

ALL
& 49.9 & -0.10 & 16.3 & 581276.38
& 50.1 & 0.04 & 14.9 & 2162.77
& \cellcolor{green!20}72.3 & \cellcolor{green!20}6.16 & \cellcolor{green!20}7.0 & \cellcolor{green!20}50.94 \\

\bottomrule
\end{tabular}%
}
\end{table}

\subsection{\texorpdfstring{\protect\circnumB{B.1}}{B.1} Supplementary Material}
\paragraph{Regularization Ablation Study}
\label{app:reg_ablation}
During the experiments, DL-specific regularization strategies were already included in the HP search: we optimized the dropout rate ({0.0, 0.2}) together with early stopping (patience = 10), see Appendix Table \ref{tab:regularization_ablation}. 
To better understand the effects of regularization on high-dimensional datasets, we additionally performed the an ablation study on a representative high-dimensional dataset (DBCD, $N=295$, $P=4912$). We evaluated DeepSurv, DeepHit, and CoxTime extending the dropout grid in total to {0.0, 0.2, 0.35, 0.5, 0.65}, selecting the best candidate among a fixed dropout rate during the search while the remaining search grid was kept unchanged.

Table~\ref{tab:regularization_ablation} presents the results by model-selection criterion. Overall, incorporating dropout into the search space is generally beneficial. In most settings, models with dropout outperform their counterparts without dropout, while the optimal dropout rate depends on the architecture and model-selection criterion.

\begin{table}[H]
\centering
\caption{Dropout ablation study on DBCD ($N=295$, $P=4912$)
for DeepHit, DeepSurv, and CoxTime, grouped by model-selection criterion. The remaining hyperparameter search grid and early stopping configuration were kept unchanged. Bold values indicate the best result within each row.}
\label{tab:regularization_ablation}
\small
\setlength{\tabcolsep}{6pt}
\begin{tabular}{lccccc}
\toprule
Method & Dropout $0.0$ & Dropout $0.2$ & Dropout $0.35$
       & Dropout $0.5$ & Dropout $0.65$ \\
\midrule
\multicolumn{6}{l}{\textbf{C-index} ($\uparrow$)} \\
\midrule
DeepHit
 & $0.653 \pm 0.077$
 & $0.699 \pm 0.046$
 & $0.703 \pm 0.082$
 & $\mathbf{0.725 \pm 0.082}$
 & $0.693 \pm 0.092$ \\
DeepSurv
 & $0.728 \pm 0.064$
 & $0.722 \pm 0.031$
 & $0.689 \pm 0.049$
 & $\mathbf{0.744 \pm 0.054}$
 & $0.715 \pm 0.048$ \\
CoxTime
 & $0.719 \pm 0.052$
 & $\mathbf{0.737 \pm 0.048}$
 & $0.735 \pm 0.050$
 & $0.699 \pm 0.075$
 & $0.726 \pm 0.055$ \\
\midrule
\multicolumn{6}{l}{\textbf{IBS} ($\downarrow$)} \\
\midrule
DeepHit
 & $0.251 \pm 0.070$
 & $0.240 \pm 0.059$
 & $0.242 \pm 0.046$
 & $0.227 \pm 0.063$
 & $\mathbf{0.226 \pm 0.008}$ \\
DeepSurv
 & $0.200 \pm 0.013$
 & $0.194 \pm 0.017$
 & $0.210 \pm 0.041$
 & $0.190 \pm 0.026$
 & $\mathbf{0.169 \pm 0.048}$ \\
CoxTime
 & $\mathbf{0.176 \pm 0.018}$
 & $0.207 \pm 0.051$
 & $0.180 \pm 0.024$
 & $0.195 \pm 0.037$
 & $0.186 \pm 0.034$ \\
\midrule
\multicolumn{6}{l}{\textbf{MAE-PO} ($\downarrow$)} \\
\midrule
DeepHit
 & $\mathbf{18.44 \pm 2.64}$
 & $18.56 \pm 2.69$
 & $18.61 \pm 2.89$
 & $18.63 \pm 2.66$
 & $18.83 \pm 2.57$ \\
DeepSurv
 & $24.44 \pm 2.78$
 & $22.63 \pm 1.98$
 & $21.46 \pm 2.16$
 & $20.59 \pm 1.84$
 & $\mathbf{19.89 \pm 0.72}$ \\
CoxTime
 & $31.45 \pm 9.26$
 & $25.93 \pm 3.57$
 & $25.07 \pm 2.82$
 & $22.36 \pm 0.84$
 & $\mathbf{19.70 \pm 1.22}$ \\
\bottomrule
\end{tabular}
\end{table}

\clearpage
\paragraph{Rank Shifts When Changing Regime}
\leavevmode \par
\begin{figure}[H]
    \centering
    \includegraphics[width=0.7\linewidth]{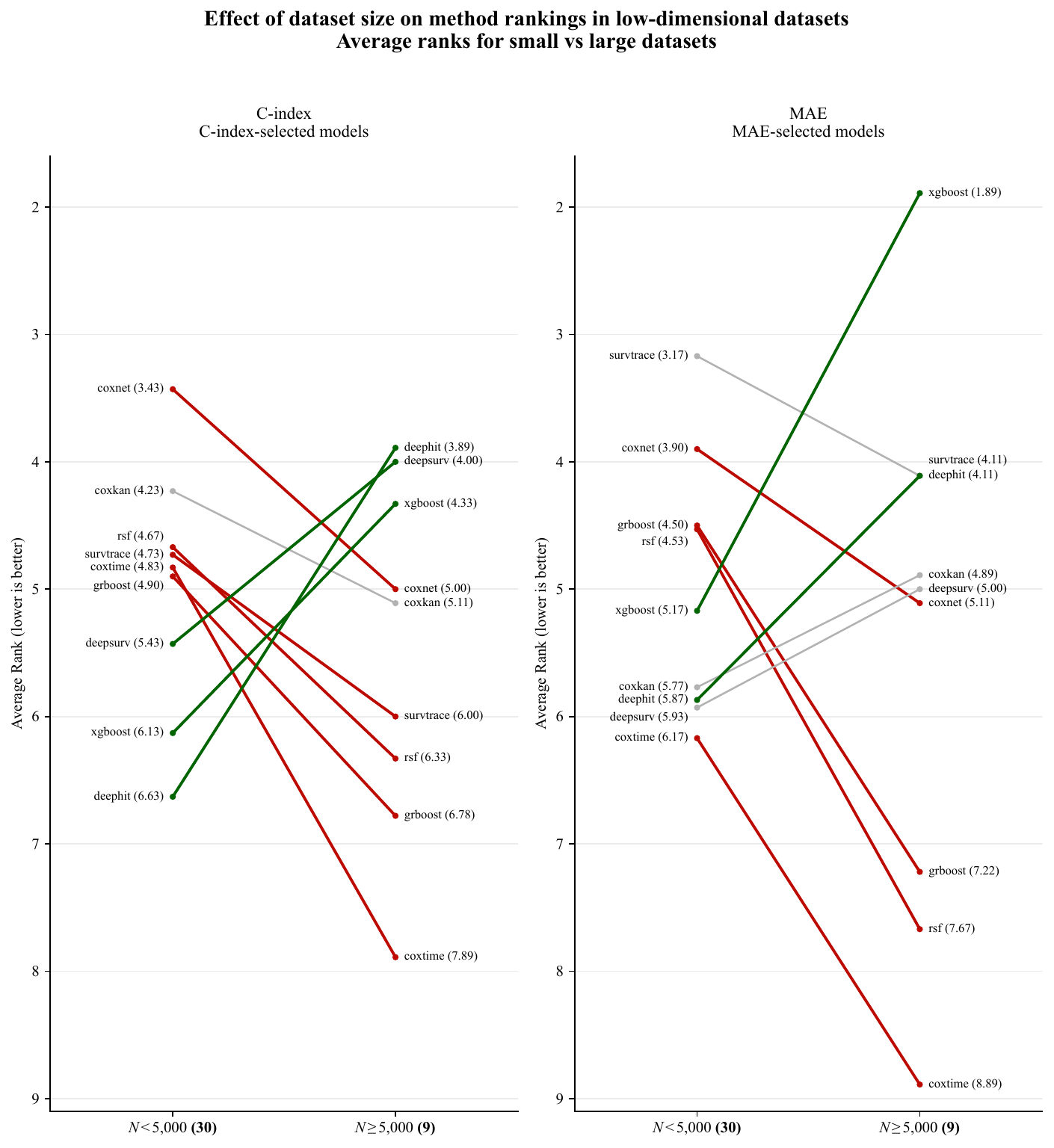}
    \caption{Effect of dataset size on method rankings in low-dimensional settings. Average ranks for smaller ($N < 5{,}000$) and larger ($N \geq 5{,}000$) datasets reveal pronounced shifts, indicating that conclusions about the best-performing methods are not stable across dataset regimes.}
    \label{fig:rank_shifts_lowdim_small_vs_large_cindex_mae}
\end{figure}

\begin{figure}[H]
    \centering
    \includegraphics[width=1\linewidth]{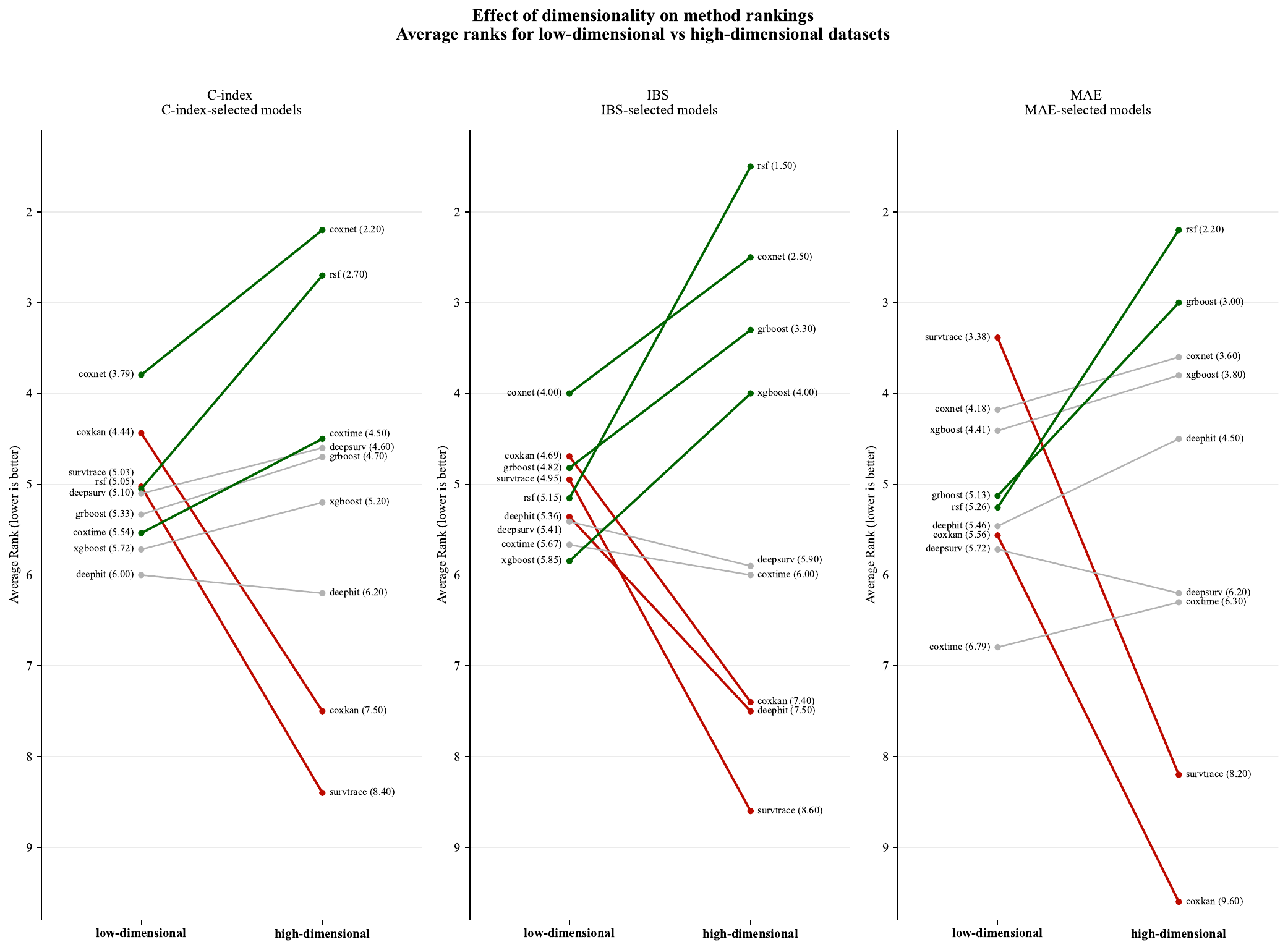}
    \caption{Effect of dimensionality on method rankings. Average ranks are shown for low-dimensional ($P \leq N$) and high-dimensional ($P > N$) datasets across three model selection settings. Substantial shifts in rankings are observed when moving to high-dimensional regimes, with tree-based methods often improving.}
    \label{fig:rank_shifts_low_vs_high_dim}
\end{figure}

\subsection{\texorpdfstring{\protect\circnumB{B.2}}{B.2}
Supplementary Material}

\paragraph{Calibration - Relative vs. Absolute Utility}
\leavevmode \par
\begin{figure}[H]
    \centering
    \includegraphics[width=0.5\linewidth]{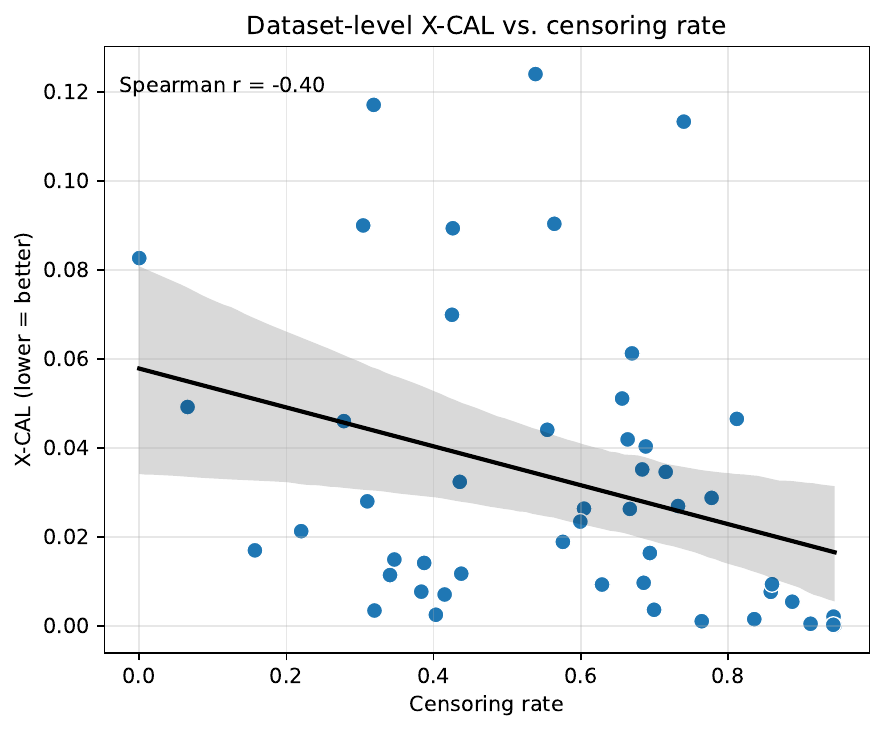}
    \caption{Dataset-level relationship between censoring rate and X-CAL. Points show the average X-CAL across all methods and model selection strategies per dataset (lower is better). A moderate negative correlation (Spearman $r=-0.40$) indicates that calibration tends to improve as censoring increases, suggesting that heavily censored datasets may lead to lower X-CAL scores.}
    \label{fig:xcal_vs_censoring_dataset_mean}
\end{figure}

\clearpage
\newpage
\input{checklist.tex}

\end{document}

%% file: checklist.tex
\section*{NeurIPS Paper Checklist}

\begin{enumerate}

\item {\bf Claims}
    \item[] Question: Do the main claims made in the abstract and introduction accurately reflect the paper's contributions and scope?
    \item[] Answer: \answerYes{}
    \item[] Justification: See sections Findings \ref{sec:findings}, and Conclusion \ref{sec:conclusion}

\item {\bf Limitations}
    \item[] Question: Does the paper discuss the limitations of the work performed by the authors?
    \item[] Answer: \answerYes{} %
    \item[] Justification: see Conclusion \ref{sec:conclusion}

\item {\bf Theory assumptions and proofs}
    \item[] Question: For each theoretical result, does the paper provide the full set of assumptions and a complete (and correct) proof?
    \item[] Answer: \answerNA{} %
    \item[] Justification: benchmarking and analysis paper; no theoretical proofs introduced or needed.

    \item {\bf Experimental result reproducibility}
    \item[] Question: Does the paper fully disclose all the information needed to reproduce the main experimental results of the paper to the extent that it affects the main claims and/or conclusions of the paper (regardless of whether the code and data are provided or not)?
    \item[] Answer: \answerYes{} %
    \item[] Justification: we describe data selection, preprocessings, used metrics, and all used libraries and reference implementations. See Section \ref{experimental_setup} and the appendix at \ref{app:dataset_preprocessing}, \ref{app:methods_and_imp}, \ref{app:eval_time_grid}. And for our GLMM, see \ref{app:GLMM} for full details.

\item {\bf Open access to data and code}
    \item[] Question: Does the paper provide open access to the data and code, with sufficient instructions to faithfully reproduce the main experimental results, as described in supplemental material?
    \item[] Answer: \answerYes{} %
    \item[] Justification: fully transparent and reproducible pipeline is provided: \url{https://github.com/sdening/surv-bench-study}.

\item {\bf Experimental setting/details}
    \item[] Question: Does the paper specify all the training and test details (e.g., data splits, hyperparameters, how they were chosen, type of optimizer) necessary to understand the results?
    \item[] Answer: \answerYes{} %
    \item[] Justification: Overview in Section \ref{experimental_setup} and all related subsections and figures. Also see the appendix at \ref{app:dataset_preprocessing}, \ref{app:methods_and_imp}, \ref{app:eval_time_grid}. And for our GLMM, see \ref{app:GLMM} for full details. All of this can be found in the provided repo link equally.

\item {\bf Experiment statistical significance}
    \item[] Question: Does the paper report error bars suitably and correctly defined or other appropriate information about the statistical significance of the experiments?
    \item[] Answer: \answerYes{} %
    \item[] Justification: We have error bars at multiple levels. We perform K-fold Cross Val for main experiments. Error bars and details are all in the appendix. For our main analysis, we model a Bayesian Generalized Linear Mixed Effects Model. We do posterior plausibility checks, fully justify all our choices. And we avoid using frequentist p-tests, as we have multiple paired ones. We stick to a fully Bayesian analysis and report Bayesian Highest Density Intervals, Effect Strengths, and Probabilities of Directions.

\item {\bf Experiments compute resources}
    \item[] Question: For each experiment, does the paper provide sufficient information on the computer resources (type of compute workers, memory, time of execution) needed to reproduce the experiments?
    \item[] Answer: \answerYes{} %
    \item[] Justification: The paper specifies the hardware and software environment used for all experiments, as well as the Python and library versions used throughout the benchmark  \ref{app:reproducibility}. All experiments were executed exclusively on CPUs without GPU acceleration. We additionally report runtime characteristics, such as the average runtime per hyperparameter configuration \ref{app:hp_grids} and the fixed compute budget used throughout the benchmark, to support reproducibility and provide transparency regarding computational requirements \ref{sec:training_protocol}.
    
\item {\bf Code of ethics}
    \item[] Question: Does the research conducted in the paper conform, in every respect, with the NeurIPS Code of Ethics \url{https://neurips.cc/public/EthicsGuidelines}?
    \item[] Answer: \answerYes{} %
    \item[] Justification: we fully adhere to the stated ethical principles.

\item {\bf Broader impacts}
    \item[] Question: Does the paper discuss both potential positive societal impacts and negative societal impacts of the work performed?
    \item[] Answer: \answerNA{} %
    \item[] Justification: a deep benchmark analysis uncovering trends in model selection strategies and dataset regimes. Has no obvious or inherent / direct links to potential harms.
    
\item {\bf Safeguards}
    \item[] Question: Does the paper describe safeguards that have been put in place for responsible release of data or models that have a high risk for misuse (e.g., pre-trained language models, image generators, or scraped datasets)?
    \item[] Answer: \answerNA{} %
    \item[] Justification: none of that applies to our paper.

\item {\bf Licenses for existing assets}
    \item[] Question: Are the creators or original owners of assets (e.g., code, data, models), used in the paper, properly credited and are the license and terms of use explicitly mentioned and properly respected?
    \item[] Answer: \answerYes{} %
    \item[] Justification: any use of code libraries or datasets is properly cited.

\item {\bf New assets}
    \item[] Question: Are new assets introduced in the paper well documented and is the documentation provided alongside the assets?
    \item[] Answer: \answerYes{} %
    \item[] Justification: We included our code base and documented it with useful comments and a sensible structure.

\item {\bf Crowdsourcing and research with human subjects}
    \item[] Question: For crowdsourcing experiments and research with human subjects, does the paper include the full text of instructions given to participants and screenshots, if applicable, as well as details about compensation (if any)? 
    \item[] Answer: \answerNA{} %
    \item[] Justification: \answerNA{}

\item {\bf Institutional review board (IRB) approvals or equivalent for research with human subjects}
    \item[] Question: Does the paper describe potential risks incurred by study participants, whether such risks were disclosed to the subjects, and whether Institutional Review Board (IRB) approvals (or an equivalent approval/review based on the requirements of your country or institution) were obtained?
    \item[] Answer: \answerNA{} %
    \item[] Justification: \answerNA{}

\item {\bf Declaration of LLM usage}
    \item[] Question: Does the paper describe the usage of LLMs if it is an important, original, or non-standard component of the core methods in this research? Note that if the LLM is used only for writing, editing, or formatting purposes and does \emph{not} impact the core methodology, scientific rigor, or originality of the research, declaration is not required.
    \item[] Answer: \answerNA{} %
    \item[] Justification: \answerNA{}

\end{enumerate}